\documentclass[10pt,twocolumn,letterpaper]{article}

\usepackage{cvpr}              %
\usepackage{subscript}
\usepackage[accsupp]{axessibility}

\usepackage[table,dvipsnames]{xcolor} 
\usepackage{multirow}

\definecolor{cvprblue}{rgb}{0.21,0.49,0.74}
\usepackage[breaklinks,colorlinks,allcolors=cvprblue]{hyperref}
\usepackage{soul}
\usepackage{graphicx}
\usepackage{lineno}
\usepackage{placeins}

\newcommand{\NSnote}[1]{\textcolor{olive}{}}

\newcommand{\NSnewnote}[1]{\textcolor{violet}{}}

\newcommand{\searchmod}{Feedback ALignment Module\xspace} %
\newcommand{\searchmodshort}{FALM\xspace} %
\newcommand{\modelname}{ReFocus}
\newcolumntype{P}[1]{>{\centering\arraybackslash}p{#1}}

\newcommand{\redtext}[1]{\textcolor{red}{#1}}
\newcommand{\greentext}[1]{\textcolor{green!60!black}{#1}}

\def\paperID{} %
\def\confName{CVPR}
\def\confYear{2026}

\title{Interactive Episodic Memory with User Feedback} 

\author{Nikesh Subedi\\
University of Utah\\
{\tt\small nikesh.subedi@utah.edu}
\and
Loris Bazzani\\
University of Verona\\
{\tt\small loris.bazzani@univr.it}
\and
Ziad Al-Halah\\
University of Utah\\
{\tt\small ziad.al-halah@utah.edu}
}

\begin{document}
\maketitle
\begin{abstract}

In episodic memory with natural language queries (EM-NLQ), a user may ask a question (e.g., ``Where did I place the mug?'') that requires searching a long egocentric video, captured from the user's perspective, to find the moment that answers it. 
However, queries can be ambiguous or incomplete, leading to incorrect responses. 
Current methods ignore this key aspect and address EM-NLQ in a one-shot setup, limiting their applicability in real-world scenarios.
In this work, we address this gap and introduce the Episodic Memory with Questions and Feedback task (EM-QnF).
Here, the user can provide feedback on the model's initial prediction or add more information (e.g., ``Before this. I'm looking for the big blue mug not the white one''), helping the model refine its predictions interactively. 
To this end, we collect datasets for feedback-based interaction and propose a lightweight training scheme that avoids expensive sequential optimization. 
We also introduce a plug-and-play \searchmod\ (\searchmodshort) that enables existing EM-NLQ models to incorporate user feedback effectively.
Our approach significantly improves over the state of the art on three challenging benchmarks and is better than or competitive with commercial large vision-language models while remaining efficient. 
Evaluation with human-generated feedback shows that it generalizes well to real-world scenarios.
Project: \url{https://nsubedi11.github.io/refocus}.
\end{abstract}

\section{Introduction}
\label{sec:intro}

\begin{figure}[t]
    \centering
    \includegraphics[width=0.9\linewidth]{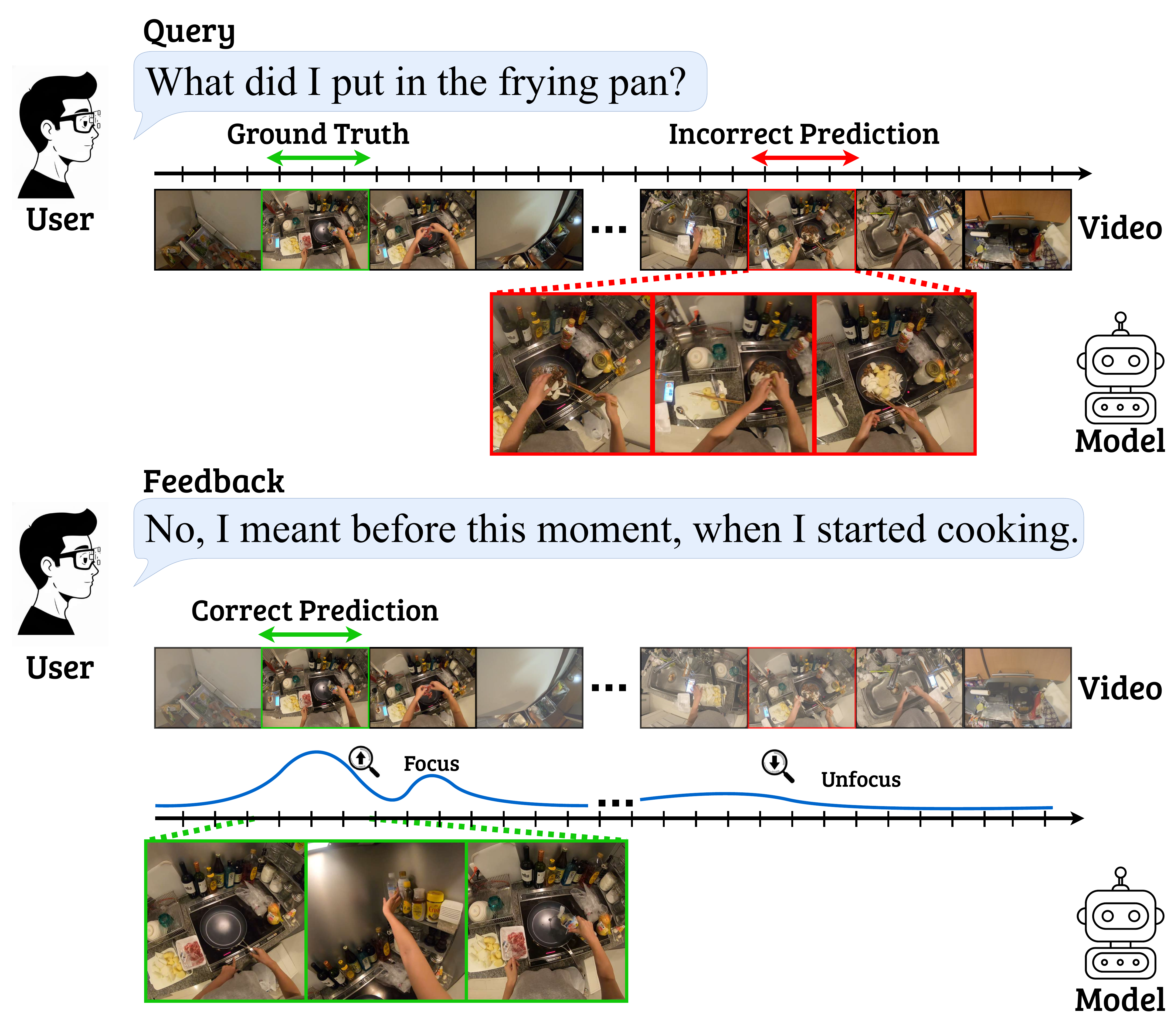}
    \caption{
        We introduce the interactive episodic memory with user feedback task (EM-QnF) to address ambiguous queries and model errors.
        Given an initial query and an incorrect model prediction (top), the user refines the query through natural language feedback, either by referring to the model's prediction or by providing additional information (middle).
        The model then uses the joint context of the query, prediction, and feedback to shift its focus toward relevant moments in the video that better align with the user's intent (bottom) to find the correct answer.
        }
        
    \label{fig:overview}
    \vspace{-0.5cm}
\end{figure}

Episodic Memory with Natural Language Query (EM-NLQ) retrieves specific moments from a person's past visual experiences, such as wearable-camera video, using free-form text questions~\cite{ego4d}. 
For example, a user might ask, ``What did I put in the frying pan?'' (Fig.~\ref{fig:overview}), and the model must identify the exact moment in the video that answers the question.  
By enabling on-demand ``visual recall,'' such systems can help users recover forgotten details, review past actions, and support embodied agents or assistive technologies in tasks such as safety checks, finding misplaced items, and retracing work steps. %

EM-NLQ remains very challenging. Egocentric videos are often long and untrimmed, making short answer segments difficult to find. 
The first-person viewpoint also introduces difficulties such as rapid head movements, motion blur, and occlusions. 
In addition, models must process long video histories efficiently while remaining lightweight enough for resource-constrained devices.

Recent work has focused on improving performance~\cite{egovlp, groundnlq, osgnet, snag, decafnet, rgnet}, efficiency~\cite{spotem, decafnet, snag}, and generalization with limited training data~\cite{naq}. 
However, a key aspect remains underexplored: user queries are often ambiguous, and real interactions are inherently iterative. 
In practice, users may refine their questions and provide feedback after seeing an incorrect prediction. 
For example, as shown in Fig.~\ref{fig:overview}, a user might respond, ``No, I meant before this moment, when I started cooking.'' %
Current EM-NLQ models cannot leverage such feedback, missing the chance to refine their predictions and better match the user's intent.

Large Vision-Language Models (LVLMs) appear to be natural candidates for addressing the interactive aspect of EM-NLQ, since they build on language models designed for dialogue, instruction following, and user alignment. 
However, in practice, current LVLMs fall short of this promise. 
As our results show, fine-tuning these models for video understanding reduces their ability to respond effectively to user feedback. 
Additionally, their reliance on large vision-language backbones makes them highly resource-intensive and slow, which limits their practicality for fast or on-device episodic memory applications.

In this work, we take a first step toward addressing the underexplored problem of interactivity and feedback in EM-NLQ.
We introduce the Episodic Memory with Questions and Feedback task (EM-QnF) and construct new datasets for feedback-based episodic memory interaction, together with a training scheme that avoids costly sequential training with feedback. 
We also propose \modelname, which integrates our novel plug-and-play \searchmod (\searchmodshort) with a variety of existing EM-NLQ models, enabling them to process and respond to user feedback efficiently and effectively. 
Unlike prior approaches that treat queries as fixed, one-shot inputs, \modelname~allows models to refine their predictions based on user corrections or clarifications, bringing EM-NLQ closer to the natural way humans seek information about past experiences.

Our approach significantly improves both the performance and scalability of EM-NLQ models in interactive settings. 
Through extensive experiments on three challenging benchmarks, our method achieves state-of-the-art performance and shows consistent gains across different EM-NLQ models and diverse evaluation settings. 
Finally, we validate the effectiveness and practicality of our approach through human-based feedback evaluations and comparisons with commercial multimodal LLMs, showing that our model can effectively incorporate real user feedback to produce more accurate and better aligned responses. %

\section{Related Works}

\paragraph{Episodic Memory with Natural Language.}
This task requires localizing a response to a query within a long egocentric video. 
Early work adapted moment localization methods to this setting, establishing strong baselines (e.g., 2D-TAN~\cite{2Dtan}, VSLNet~\cite{vslnet}). 
Subsequent methods incorporated multiscale representations to handle the wide variation in response durations~\cite{groundnlq} or used object-centric features to capture a broader range of objects~\cite{osgnet}. 
To address the scarcity of labeled data, some approaches leveraged more readily available supervision such as narrations~\cite{naq} or adapted video-text contrastive learning to egocentric videos~\cite{egovlp}. 
Other works~\cite{snag, decafnet, spotem, rgnet} focused on improving the efficiency and accuracy of EM-NLQ models.

However, all previous EM-NLQ methods localize the query in a one-shot manner and therefore cannot handle query ambiguity or model errors. In contrast, we propose an interactive episodic memory search setting that allows users to provide feedback to help the model refine its prediction and identify the correct response.

\vspace{-0.4cm}
\paragraph{Large Vision-Language Models.}

Large Vision-Language Models (LVLMs)~\cite{llava,vid-llama, internvl, blip, blip2, qwen25vl} possess broad visual knowledge and excel at visual-linguistic reasoning and spatial understanding.
However, these methods often struggle with time-sensitive video tasks such as temporal localization, especially in long videos. 
Many works~\cite{sevila,llavilo,vtimellm,timechat,timemarker,lita} have proposed solutions to improve the alignment between video semantics and their corresponding timestamps for such tasks. 
For example, TimeChat~\cite{timechat} introduces timestamp-aware video representations for temporal grounding, UniTime~\cite{unitime} proposes a multi-stage inference strategy for moment localization in long videos, and ChatVTG~\cite{chatvtg} explores training-free temporal grounding with LVLMs. 

Despite strong performance on some of these tasks, these methods remain computationally expensive and, as we show in our experiments, fail to generalize well or retain their instruction-following ability when adapting to user feedback. 
Instead, we propose a plug-and-play feedback alignment module that enables EM-NLQ models to effectively use user feedback, leading to significant performance improvements while keeping computational cost low.

\vspace{-0.4cm}
\paragraph{Localization with Textual Feedback. }
Utilizing human or synthetically generated feedback has been shown to effectively improve model capabilities across language reasoning~\cite{madaan2023self, wu2023fine, ziegler2019fine} and visual understanding~\cite{li-etal-2024-vlfeedback}. 
Different forms of feedback have been explored, including regions for segmentation~\cite{wei2023focused}, clicks for preference learning~\cite{ziegler2019fine}, and timestamps for temporal grounding~\cite{dong2024temporal,unitime}. 
Among these, language feedback remains one of the richest and most user-friendly ways to refine predictions at inference time~\cite{guo2018dialog, maeoki2020interactive, wu2021fashion, levy2023chatting} and to finetune models~\cite{wu2023fine}. 
Closely related work also appears in localization within navigational environments~\cite{hahn2020you, thomason2020vision}, where a locator tries to find an observer in an indoor environment, and the observer provides natural language feedback to guide the search.

However, to the best of our knowledge, textual feedback has not yet been shown to be effective for EM-NLQ. 
In this work, we propose an approach that leverages textual feedback to resolve localization ambiguities caused by large search spaces and imprecise user queries in egocentric videos. 
We further propose a feedback generation recipe that enables training at scale without expensive manual annotation, while still generalizing well to human feedback.

\section{Episodic Memory with User Feedback}

Our work is the first to explore the interactive aspect of EM-NLQ by enabling models to refine their predictions based on user feedback. 
We first formally define this new task (Sec.~\ref{ssec:task}), then describe an effective procedure for generating feedback data for training models in this setting (Sec.~\ref{ssec:recipe-data}). 
Finally, we introduce \searchmod~(\searchmodshort), a plug-and-play module that can be seamlessly integrated into existing EM-NLQ models, allowing them to process and respond to user feedback (Sec.~\ref{ssec:pnp-module}).

\subsection{Task Definition}\label{ssec:task}

We introduce the \textbf{Episodic Memory with Questions and Feedback} task (EM-QnF). 
Unlike EM-NLQ, which answers a natural language query in a one-shot manner, EM-QnF extends the task by allowing users to provide natural language feedback that guides the model toward a more accurate prediction. 
The goal is to enable EM-NLQ models to iteratively refine their responses based on user feedback and prior outputs.

Formally, given an egocentric video $\mathcal{V}$ and a natural language query $\mathcal{Q}$, the objective is to identify the video segment $\mathcal{R}^q \in \mathcal{V}$ that answers $\mathcal{Q}$, where $\mathcal{R} = [t_s, t_e]$ denotes the temporal window of the response defined by its start and end times. 
An EM-NLQ model produces an initial prediction $\mathcal{R}_1$, which may be incorrect due to query ambiguity, missing context, or model error. 
The user then provides feedback $\mathcal{F}_1$, a natural language statement containing additional or contrastive information relative to $\mathcal{R}_1$. 
The model integrates $\mathcal{F}_1$ in the context of $(\mathcal{V}, \mathcal{Q}, \mathcal{R}_1)$ to generate a refined prediction $\mathcal{R}_2 \neq \mathcal{R}_1$. 
This interactive process continues over multiple rounds, producing a sequence of responses $\{\mathcal{R}_1, \dots, \mathcal{R}_n\}$ and feedbacks $\{\mathcal{F}_1, \dots, \mathcal{F}_n\}$ until the final response $\mathcal{R}_n$ matches the correct answer, $\mathcal{R}_n = \mathcal{R}^q$. 
Without loss of generality, we refer to the current model prediction at step i as the \emph{reference span} $\mathcal{R}^f$, which the user provides feedback on, and focus on the single-turn case in the following sections.
We provide an extension to the multi-turn scenario in Sec. \ref{ssec:pnp-module}.

\begin{figure}[t]
    \centering
    \includegraphics[width=0.99
    \linewidth]{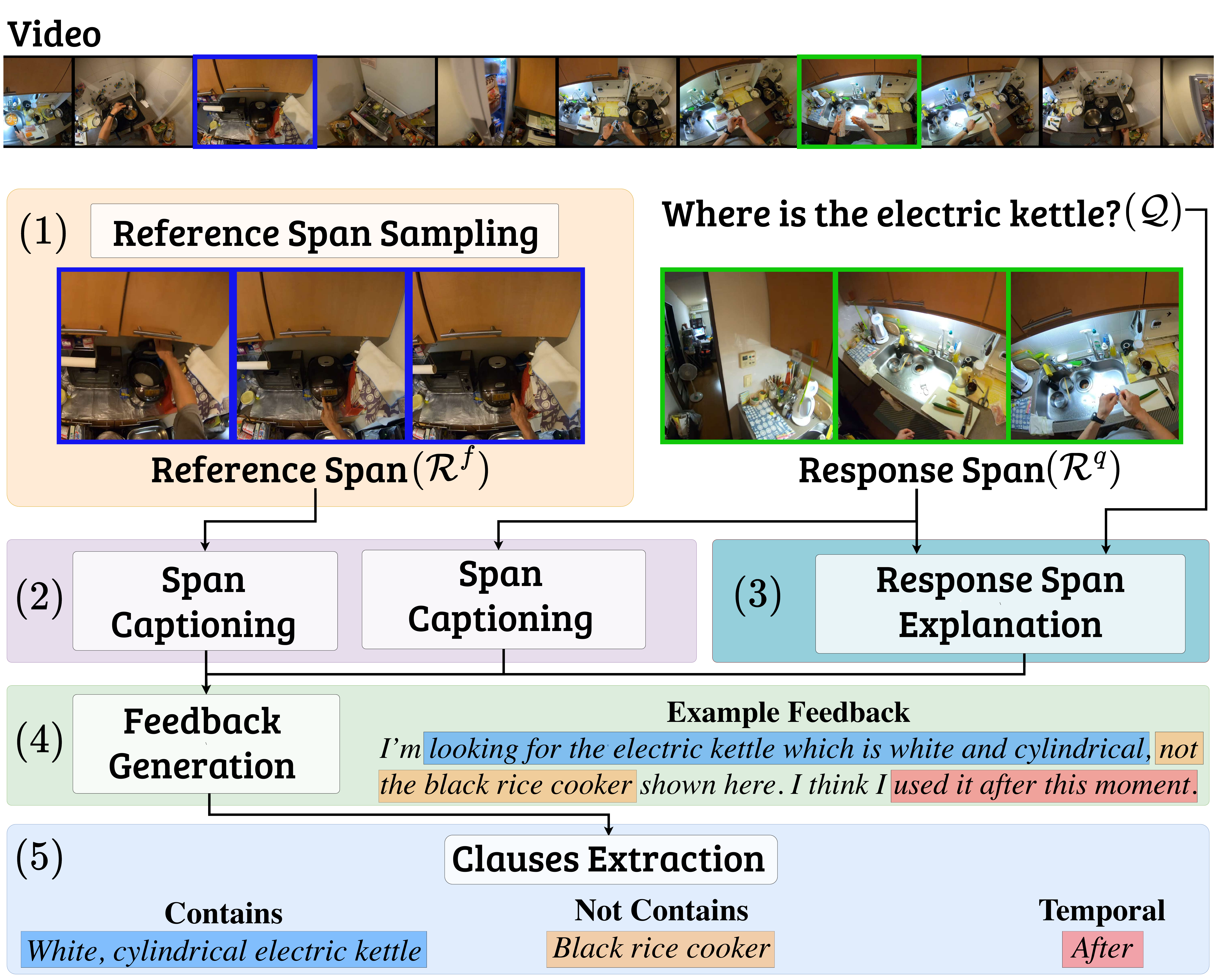}
    \caption{{Our feedback generation recipe. For a query $\mathcal{Q}$ and ground-truth response span $\mathcal{R}^q$, (1) we sample a reference span $\mathcal{R}^f$, then (2) collect captions describing each span. 
    Additionally, (3) we collect an explanation of why $\mathcal{R}^q$ answers $\mathcal{Q}$. 
    The captions from (2) and (3) are then used to generate a feedback $\mathcal{F}$. 
    Finally, (5) we extract three \textit{clauses} from $\mathcal{F}$ representing different types of information, which we use to generate labels that supervise the learning of our \searchmodshort module (see Sec.~\ref{ssec:pnp-module}).}
        }
    \label{fig:overview_clauses}
    \vspace{-0.5cm}
\end{figure}

\begin{figure*}[t]
    \centering
    \includegraphics[width=0.99\linewidth]{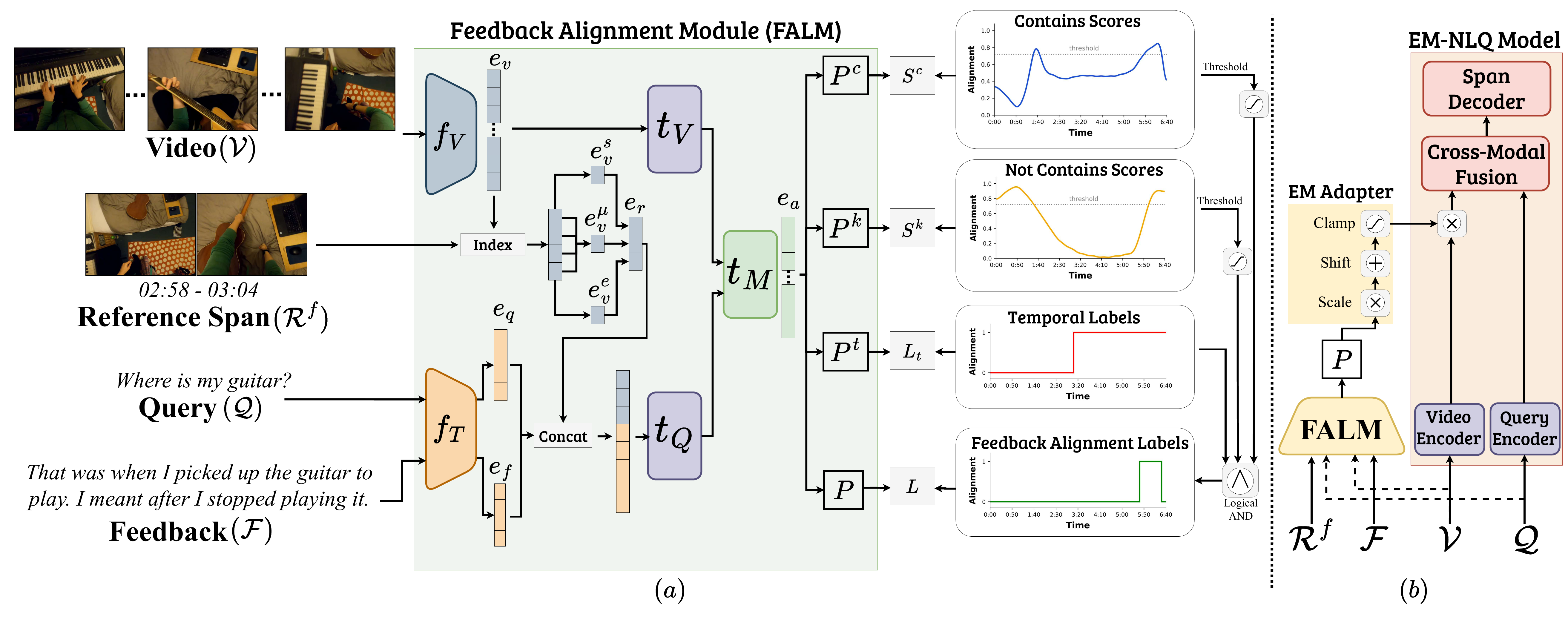}
    \caption{ 
        Our (a) \searchmod~(FALM) module and (b) its integration with an EM-NLQ model, ReFocus.
        FALM is trained to predict an alignment score $P$ that indicates how well each clip aligns with the user feedback $\mathcal{F}$ in the context of the input video $\mathcal{V}$, the query $\mathcal{Q}$, and the reference span $\mathcal{R}^f$.
        It is then plugged into an EM-NLQ model using a lightweight adapter (b), enabling the model to leverage user feedback effectively by shifting its focus toward video clips that better match the user intent expressed in the feedback.
            }
    
    \label{fig:overview_model}
\end{figure*}

\subsection{A Recipe for User Feedback Generation}
\label{ssec:recipe-data}

To the best of our knowledge, no publicly available dataset currently supports the EM-QnF task. 
Collecting real user feedback is costly and time-consuming, as it requires human annotators to watch long egocentric videos and write meaningful feedback for incorrect model predictions. 

To address this limitation, we propose a \emph{synthetic feedback generation recipe} that produces useful and realistic feedback data from existing EM-NLQ datasets, effectively turning them into EM-QnF datasets. 
Our recipe, shown in Fig.~\ref{fig:overview_clauses}, has four main steps: (1) reference span sampling, (2) caption generation for the sampled clips, (3) explanation generation for response span and (4) feedback construction. 
This approach enables scalable and controlled generation of feedback examples without costly manual annotation. 
Furthermore, our results show that models trained on the proposed synthetic feedback data can effectively use real user feedback at inference time. 
Notably, synthetic feedback yields improvements comparable to those obtained with real user feedback (see Sec.~\ref{ssec:human-eval}), suggesting that our feedback generation recipe produces realistic feedback that aligns well with human feedback styles.
Next, we describe the main steps of the proposed recipe, and provide more details in the supplementary material.

\vspace{-0.3cm}
\paragraph{Reference Span Sampling.}
The goal of this step is to sample a span $\mathcal{R}^f$ that simulates an incorrect prediction for a given query $\mathcal{Q}$. 
An intuitive approach is to use actual model failures from an EM-NLQ model as reference spans. 
However, relying only on such examples limits the diversity of error types and may lead to overfitting to the behavior of a specific model. 
Hence, we sample two additional types of spans: (1) $\mathcal{R}^q$-similar spans, which are visually similar to the ground-truth response $\mathcal{R}^q$ (based on video feature similarity) but do not correctly answer the query ($\mathcal{R}^f \neq \mathcal{R}^q$); and (2) random spans, which are randomly sampled segments that are temporally disjoint from $\mathcal{R}^q$. 
We refer to spans sampled from model failures or $\mathcal{R}^q$-similar spans as \emph{query-relevant spans}, since they may contain some information relevant to the query but are still incorrect. In contrast, we refer to random spans as \emph{query-irrelevant spans}, since they represent completely off-target responses.

\vspace{-0.3cm}
\paragraph{Response Captioning and Explanation Generation.}
For all ground-truth response spans $\mathcal{R}^q$ and the sampled reference spans $\mathcal{R}^f$ from previous step, we first generate textual captions describing these spans.  
This step reduces the computational cost of feedback generation by removing the need to process long video clips directly with large vision-language models (LVLMs), and it improves feedback quality by providing concise and relevant summaries that large language models (LLMs) can reason over effectively. 
For each span $\mathcal{R}_i$, we use a pre-trained LVLM to generate a textual description $\mathcal{D}_i$ that captures the visual content (e.g., objects and scenes) and actions present in the span, independent of any query $\mathcal{Q}$. 
Additionally, for each ground-truth span $\mathcal{R}^q_i$, we generate an explanation $E_i$ describing why the span answers its corresponding query $\mathcal{Q}_i$. This helps us control the type of information allowed in the feedback, as explained next.

\vspace{-0.3cm}
\paragraph{Feedback Generation.}
The final step generates natural language feedback $\mathcal{F}$ that guides the model from a sampled reference span $\mathcal{R}^f$ toward the correct ground-truth span $\mathcal{R}^q$.  
For each query $\mathcal{Q}_i$, we sample pairs $(\mathcal{R}^q_i, \mathcal{R}^f_j)$ and provide their corresponding descriptions $\mathcal{D}^q_i$, $\mathcal{D}^f_j$, the explanation $E^q_i$, and the relative temporal order between $\mathcal{R}^q_i$ and $\mathcal{R}^f_j$ to a reasoning-focused large language model (LLM). 
The LLM is prompted to generate feedback $\mathcal{F}_{i,j}$ that provides informative cues to help locate $\mathcal{R}^q_i$ without directly revealing the answer to $\mathcal{Q}_i$.
To achieve this, we design prompts that encourage the feedback to include any combination of three types of information:  
(1) additional disambiguating details about the queried object or moment derived from $\mathcal{D}^q_i$;  
(2) contrastive cues highlighting differences between $\mathcal{R}^q_i$ and $\mathcal{R}^f_j$ based on $\mathcal{D}^q_i$ and $\mathcal{D}^f_j$; and  
(3) temporal guidance indicating whether to search before or after the current reference span $\mathcal{R}^f_j$.  
We further use the explanation $E^q_i$ to instruct the LLM not to produce feedback that trivially answers the original query $\mathcal{Q}_i$.

This leads to diverse types of feedback, ranging from short phrases (e.g., ``before this'') that simulate impatient users to more descriptive feedback that may include different types of information based on the cues above. The feedback has an average length of 16 words, with a standard deviation of 6.8 (see Supp for the prompts and examples).

\vspace{-0.3cm}
\paragraph{EM-QnF Samples.}
Following this recipe, the constructed EM-QnF dataset consists of samples of the form  
$(\mathcal{V}_i, \mathcal{Q}_i, \mathcal{R}^q_i, \{(\mathcal{R}^f_{i,j}, \mathcal{F}_{i,j})\})$,  
which can be used to train models to handle interactive feedback refinement.

\subsection{\searchmodshort: \searchmod}\label{ssec:pnp-module}

We introduce \searchmodshort, a plug-and-play module designed to help EM-NLQ models align with and use user feedback.
Given user feedback $\mathcal{F}$, a reference span $\mathcal{R}^f$, a video $\mathcal{V}$, and a natural language query $\mathcal{Q}$, the core idea of \searchmodshort~is to predict an alignment score for each clip in $\mathcal{V}$, indicating its relevance to $\mathcal{F}$. %

Modern video encoders typically process long videos by dividing them into short clips and encoding each clip independently before aggregation. 
Formally, let $\mathcal{V} = \{C_1, \dots, C_m\}$ denote a video segmented into $m$ clips, where $C_i$ is the $i$-th clip. 
\searchmodshort~takes $(\mathcal{V}, \mathcal{Q}, \mathcal{R}^f, \mathcal{F})$ as input and outputs an alignment vector $P \in [0,1]^m$, 
where each element $P_i$ represents the alignment score between clip $C_i$ and the given user feedback $\mathcal{F}$. 
These scores are then used to reweight the video features in existing EM-NLQ models, effectively shifting the model's attention toward clips that are most relevant to the user feedback when predicting the next response span.

\vspace{-0.3cm}
\paragraph{\searchmodshort Architecture.} 
As shown in Fig.~\ref{fig:overview_model}.a, \searchmodshort encodes the video $\mathcal{V}$, query $\mathcal{Q}$, and feedback $\mathcal{F}$ using pretrained video and text encoders, $f_V$ (ViT\textendash1B from EgoVideo~\cite{egovideo}) and $f_T$ (gte-Qwen2-7B-instruct~\cite{qwen_gte}), respectively. 
This yields feature representations $f_V(\mathcal{V}) = e_v \in \mathbb{R}^{v_t \times d}$, $f_T(\mathcal{Q}) = e_q \in \mathbb{R}^{q_t \times d}$, and $f_T(\mathcal{F}) = e_f \in \mathbb{R}^{f_t \times d}$, 
where $d$ is the model dimension and $(v_t, q_t, f_t)$ denote the number of tokens in each input sequence. 
We represent the reference span $\mathcal{R}^f$ by concatenating its start, end, and mean clip embeddings from the video features, 
$e_r = [e_v^s, e_v^e, e_v^\mu] \in \mathbb{R}^{3 \times d}$, 
allowing \searchmodshort to interpret the feedback $\mathcal{F}$ in the visual context of the reference span. 

Next, we use a two-layer Transformer encoder ($t_Q$) to model interactions among $\{e_q, e_f, e_r\}$ and another two-layer Transformer encoder ($t_V$) to capture the overall video context from $e_v$. 
Finally, we pass the features through two Transformer decoder blocks with cross-attention ($t_M$) to produce video-feedback aligned embeddings $e_a$.

\vspace{-0.3cm}
\paragraph{Alignment Supervision.} %
We generate pseudo-labels to supervise the learning of \searchmodshort by indicating clip relevance based on three cues extracted from the feedback: 
(1) Contains: what information the correct response should contain, 
(2) Not Contains: what should not appear in the response, and 
(3) Temporal: whether to search before or after the reference span.

We prompt an LLM to extract these cues from each feedback $\mathcal{F}$ in form of short language clauses (see Fig.~\ref{fig:overview_clauses}(5)). 
Using the EgoVideo~\cite{egovideo} encoders, we compute \textless clip, clause\textgreater\ similarities to obtain \emph{contains} scores $S^c$ and \emph{not-contains} scores $S^n$.
To reduce noise in these similarity measures, we apply Gaussian smoothing, and min-max normalization across all clips.
Finally, we invert the $S^n$ scores to obtain $S^k = 1 - S^n$, i.e. the model should avoid clips that include information the user excluded in their feedback. 
To convert these scores into binary labels, we first calculate the mean and standard deviation of the scores within the correct response $\mathcal{R}^q$~\cite{sgdetr}, denoted by $S_\mu$ and $S_\sigma$ for each score type, $S^c$ and $S^k$. 
The threshold is then defined as $\delta = S_\mu - 3S_\sigma$. 
We assign a label of 1 to clips with scores above the threshold and 0 otherwise, resulting in binary labels $L^c$ and $L^k$ based on $\delta^c$ and $\delta^k$. 
For temporal labels $L^t$, we assign 1 to clips before or after $\mathcal{R}^f$ according to the extracted temporal clause. 
Finally, we combine the three label types using a logical AND operation to form the final \searchmodshort labels: $L = L^c \land L^k \land L^t$. 
Note that not all three clauses are present in every feedback instance, and the labels are generated using only the subset of clauses extracted from each $\mathcal{F}$.

\vspace{-0.3cm}
\paragraph{\searchmodshort Training Objective.}
Given encoded features $e_a$, four MLP heads predict \emph{contains} ($P^c$) \emph{not-contains} ($P^k$), \emph{temporal} ($P^t$), and overall alignment ($P$) scores:
\begin{align}
\mathcal{L} =&
\lambda\mathcal{L}_{C}(L, P) +
\lambda_t\mathcal{L}_{C}(L^t, P^t) +\\
& \lambda_c\mathcal{L}_{2}(S^c, P^c) +
\lambda_n\mathcal{L}_{2}(S^k, P^k),
\end{align}

where $\mathcal{L}_{C}$ is a binary cross-entropy loss and $\mathcal{L}_{2}$ is $\|\cdot\|_2^2$ regression loss. 

\vspace{-0.3cm}
\paragraph{\modelname: \searchmodshort Integration.}
After pretraining \searchmodshort on feedback data, we can integrate it into an EM-NLQ model by reweighting that model's video clip features using the predicted alignment scores $P$ as shown in Fig.~\ref{fig:overview_model}(b). 
In other words, \searchmodshort shifts the focus of the EM-NLQ model by emphasizing or de-emphasizing the importance of certain clips based on user feedback.
To enable effective and seamless adaptation across EM-NLQ models, we introduce a lightweight EM Adapter that scales and shifts the alignment scores of \searchmodshort using two learned scalars, $\alpha$ and $\beta$, while fine-tuning with the EM-NLQ model:
$\hat{P} = \text{clamp}(\alpha P + \beta, 0, 1)$. 
We denote the final approach with \searchmodshort plugged into an EM-NLQ model $\mathcal{M}$ with the adapter as \modelname($\mathcal{M}$).

\vspace{-0.3cm}
\paragraph{Multi-Turn Feedback Extension.}
While our work focuses on single-turn feedback, we propose here an extension of \modelname~to the multi-turn feedback setting. 
Given multiple independent feedback samples $\{\mathcal{F}_1, \dots, \mathcal{F}_n\}$ for the same query $\mathcal{Q}$, we first pass each feedback to \modelname($\mathcal{M}$) and use late fusion by averaging the output features of the Cross-Modal Encoder obtained from integrating each feedback separately, before passing the fused features to the Span Decoder, as shown in Fig.~\ref{fig:overview_model}(b), to localize the answer. 
We find that this simple extension leads to significant gains without added complexity (Sec.~\ref{ssec:model_analysis}). 
More advanced multi-turn modeling remains an important direction for future work, and our extension provides a strong baseline.

\section{Experiments}

\begin{table*}[ht]
\centering
\small
\setlength{\tabcolsep}{2.9pt} 
\begin{tabular}{ll|cccc|cccc|cccc@{}}
\toprule
& &
\multicolumn{4}{c|}{\textbf{\fontsize{9.5}{13}\selectfont Ego4D-QnF}} &
\multicolumn{4}{c|}{\textbf{\fontsize{9.5}{13}\selectfont GoalStep-QnF}} &
\multicolumn{4}{c}{\textbf{\fontsize{9.5}{13}\selectfont HD-EPIC-QnF}} \\[2pt]
 & & \multicolumn{2}{c}{\textbf{IoU = 0.3}} & \multicolumn{2}{c|}{\textbf{IoU = 0.5}}
 & \multicolumn{2}{c}{\textbf{IoU = 0.3}} & \multicolumn{2}{c|}{\textbf{IoU = 0.5}}
 & \multicolumn{2}{c}{\textbf{IoU = 0.3}} & \multicolumn{2}{c}{\textbf{IoU = 0.5}} \\[2pt]
 & \textbf{\fontsize{9.5}{13}\selectfont Method} & \textbf{R1} & \textbf{R5} & \textbf{R1} & \textbf{R5} & \textbf{R1} & \textbf{R5} & \textbf{R1} & \textbf{R5} & \textbf{R1} & \textbf{R5} & \textbf{R1} & \textbf{R5} \\
\midrule

\multirow{3}{*}{\rotatebox[origin=c]{90}{\textbf{LVLM}}} &
TimeChat (ZS) & 1.6\redtext{\textsuperscript{-0.2}} & N/A & 0.7\redtext{\textsuperscript{-0.2}} & N/A & 2.3\greentext{\textsuperscript{+0.9}} & N/A & 1.1\greentext{\textsuperscript{+0.7}} & N/A & 0.2\redtext{\textsuperscript{+0.0}} & N/A & 0.0\redtext{\textsuperscript{+0.0}} & N/A \\[2pt]

& UniTime (ZS) & 19.9\redtext{\textsuperscript{-5.2}} & N/A & 12.3\redtext{\textsuperscript{-3.3}} & N/A & 10.2\redtext{\textsuperscript{-1.7}} & N/A & 6.0\redtext{\textsuperscript{-0.8}} & N/A & 3.6\redtext{\textsuperscript{-2.0}} & N/A & 1.3\redtext{\textsuperscript{-1.2}} & N/A \\[2pt]

& UniTime (FT) & 21.7\redtext{\textsuperscript{-3.4}} & N/A & 13.3\redtext{\textsuperscript{-2.3}} & N/A & 8.2\redtext{\textsuperscript{-0.3}} & N/A & 5.5\redtext{\textsuperscript{-0.1}} & N/A & 2.5\redtext{\textsuperscript{-1.6}} & N/A & 1.0\redtext{\textsuperscript{-0.8}} & N/A \\[2pt]
\midrule

\multirow{4}{*}{\rotatebox[origin=c]{90}{\textbf{Task-Expert}}} &
{OSGNet} & 29.6\greentext{\textsuperscript{+0.4}} &	56.3\greentext{\textsuperscript{+0.5}} &	20.5\greentext{\textsuperscript{+0.5}} &	43.3\greentext{\textsuperscript{+0.7}} &	30.2\greentext{\textsuperscript{+0.6}} &	60.1\greentext{\textsuperscript{+0.9}} &	24.7\greentext{\textsuperscript{+0.5}} &	52.5\greentext{\textsuperscript{+0.7}} &	14.7\greentext{\textsuperscript{+0.3}} &	37.7\redtext{\textsuperscript{-0.1}} &	9.6\greentext{\textsuperscript{+0.1}} &	25.2\greentext{\textsuperscript{+0.1}} \\		

& {ReFocus(OSGNet)} & \textbf{32.5\greentext{\textsuperscript{+3.3}}} &	\textbf{58.3\greentext{\textsuperscript{+3.7}}} &	\textbf{22.4\greentext{\textsuperscript{+1.9}}} &	\textbf{45.3\greentext{\textsuperscript{+3.3}}} &	\textbf{31.9\greentext{\textsuperscript{+2.0}}} &	\textbf{61.0\greentext{\textsuperscript{+2.3}}} &	\textbf{26.5\greentext{\textsuperscript{+1.8}}} &	\textbf{53.7\greentext{\textsuperscript{+2.4}}} &	\textbf{15.3\greentext{\textsuperscript{+0.8}}} &	\textbf{38.3\greentext{\textsuperscript{+1.3}}} &	\textbf{10.1\greentext{\textsuperscript{+0.5}}} &	\textbf{25.8\greentext{\textsuperscript{+0.9}}} \\	
\cmidrule{2-14}
& GroundNLQ & 29.6\greentext{\textsuperscript{+0.6}} & 56.0\greentext{\textsuperscript{+1.0}} & 21.4\greentext{\textsuperscript{+0.2}} & 43.0\greentext{\textsuperscript{+0.6}} & 23.3\greentext{\textsuperscript{+0.2}} & 53.2\greentext{\textsuperscript{+0.3}} & 17.9\greentext{\textsuperscript{+0.5}} & 43.7\greentext{\textsuperscript{+0.4}} & 11.3\redtext{\textsuperscript{+0.0}} & 33.8\greentext{\textsuperscript{+0.9}} & 6.5\redtext{\textsuperscript{-0.1}} & 21.1\greentext{\textsuperscript{+0.5}} \\[2pt]

& ReFocus(GroundNLQ) & \textbf{33.1\greentext{\textsuperscript{+3.3}}} &\textbf{59.7\greentext{\textsuperscript{+4.6}}} & \textbf{23.7\greentext{\textsuperscript{+2.2}}} & \textbf{46.1\greentext{\textsuperscript{+3.9}}} & \textbf{26.8\greentext{\textsuperscript{+4.9}}} & \textbf{56.2\greentext{\textsuperscript{+5.4}}} & \textbf{20.3\greentext{\textsuperscript{+3.6}}} & \textbf{46.1\greentext{\textsuperscript{+4.8}}} & \textbf{15.1\greentext{\textsuperscript{+3.0}}} & \textbf{39.6\greentext{\textsuperscript{+5.4}}} & \textbf{9.1\greentext{\textsuperscript{+2.1}}} & \textbf{25.7\greentext{\textsuperscript{+4.0}}} \\

\bottomrule
\end{tabular}
\caption{Model performance comparison across QnF datasets. Deltas ($\Delta$) of feedback vs query-only performance are shown as superscripts. For LVLMs, {ZS denotes zero-shot evaluation of the method, while FT is after finetuning on QnF data.} 
}
\label{tab:combined_feedback_results}
\end{table*}

We demonstrate the effectiveness of our approach on three challenging egocentric video datasets adapted to the EM-QnF task (Sec.~\ref{ssec:eval_setup}), and compare it against state-of-the-art EM-NLQ and LVLM models (Sec.~\ref{ssec:main_results}). 
We then provide an analysis of our model (Sec.~\ref{ssec:model_analysis}), including comparisons with commercial LVLMs (Gemini-Flash) and human-generated feedback. 
Finally, we show the performance of our approach in multi-turn feedback scenarios.

\subsection{Evaluation Setup}
\label{ssec:eval_setup}
\paragraph{Datasets.} 
We experiment with Ego4D-NLQ~\cite{ego4d}, Ego4D-GoalStep~\cite{goalstep}, and HD-EPIC~\cite{hdepic}, which are widely used egocentric video benchmarks. 
While Ego4D-NLQ already provides query and response annotations, GoalStep and HD-EPIC are not designed for the NLQ task. 
Using NLQ templates from Ego4D-NLQ, we leverage the step descriptions from GoalStep and the narrations from HD-EPIC to generate natural language queries. 
We then apply our feedback generation recipe to all three datasets, resulting in question-and-feedback datasets that we refer to as Ego4D-QnF, GoalStep-QnF, and HD-EPIC-QnF. 
See Supp for dataset details and statistics.

\vspace{-0.3cm}
\paragraph{Comparison with the SoTA.} %
We compare \modelname\ against several SoTA methods: LVLM-based methods like \textbf{TimeChat}~\cite{timechat} and \textbf{UniTime}~\cite{unitime}, and the task expert EM-NLQ models like \textbf{GroundNLQ}~\cite{groundnlq} and \textbf{OSGNet}~\cite{osgnet}. 
We evaluate UniTime and TimeChat in zero-shot setting and additionally, finetune UniTime on EM-QnF datasets as well. 
For task experts, we first pretrain on NaQ~\cite{naq} and Ego4D NLQ~\cite{ego4d} datasets using EgoVideo~\cite{egovideo} as video features and gte-Qwen2-7B-instruct~\cite{qwen_gte} as text features.
We adapt EM-NLQ models to feedback input by concatenating the text features with reference span and feedback features to form a new query representation.
Our \searchmodshort module is applied to GroundNLQ and OSGNet, which are named \modelname(GroundNLQ) and \modelname(OSGNet). 
We train all models (with and without our module) on the same QnF data for fair comparisons.

\vspace{-0.5cm}
\paragraph{Implementation Details.} 
Videos are divided into non-overlapping clips of 16 consecutive frames each. %
The clips are encoded with $f_V=$ ViT\textendash1B from EgoVideo~\cite{egovideo}. 
For text features, we use $f_T=$ gte-Qwen2-7B-instruct~\cite{qwen_gte}. 
We pretrain \searchmodshort on the combined training splits of the three QnF datasets. 
For Refocus, we integrate \searchmodshort into the EM-NLQ model pretrained on NaQ~\cite{naq} and fine-tune Refocus($\mathcal{M}$) on the combined training splits of the three QnF datasets. See Supp for full details.

\vspace{-0.5cm}
\paragraph{Evaluation Metrics.} Following the episodic memory benchmark~\cite{ego4d}, we report Recall (R1 and R5) at multiple temporal intersection-over-union (tIoU) thresholds \{0.3, 0.5\} between the predicted and ground-truth spans.

\subsection{Main Results}
\label{ssec:main_results}

Table~\ref{tab:combined_feedback_results} shows the performance of our model and the baselines when evaluated with feedback.  
For brevity, we report results in the format $X^{\Delta}_{q+f}$, where $\Delta = X_{q+f} - X_q$, $X_{q+f}$ is the model performance when given both the query and the feedback, and $X_q$ is its performance when given only the query (i.e., the initial performance before feedback).  
Thus, $X_{q+f}$ shows the absolute performance after processing the feedback, while $\Delta$ shows whether, and by how much, the model benefits from the feedback.

Interestingly, the LVLM-based localization methods (Table~\ref{tab:combined_feedback_results}, top), which incorporate an LLM in their architecture, fail to adapt to user feedback. In most metrics and across datasets, they perform worse with feedback than without it, resulting in negative $\Delta$ values. 
Even when fine-tuned on the new QnF data, these models do not show clear improvement in leveraging feedback, despite the stronger reasoning capabilities of their underlying LLMs, which might be expected to help them better handle such interactions.

As for EM-NLQ experts, simply training OSGNet and GroundNLQ on EM-QnF data leads to only small improvements, with $\Delta \leq 1\%$ across all metrics. 
Although these models are trained with EM-QnF data as well, they still largely ignore the feedback. 

Across all datasets, our \modelname~consistently outperforms all other methods. 
Specifically, \modelname(GroundNLQ) achieves notable gains on R1 and R5, with up to +4.9 and +5.4, respectively. 
We also observe consistent improvements when \modelname~is applied to OSGNet. 
These results demonstrate the effectiveness of our approach in leveraging user feedback to improve localization across EM-NLQ models and benchmarks. 
Furthermore, our approach does not rely on heavyweight components such as LLMs and preserves the efficiency of the underlying task-expert model (see Supp for details).

\begin{table}[t]
\centering
\small
\setlength{\tabcolsep}{3.5pt} 
\begin{tabular}{l|cc|cc}
\toprule
 &
\multicolumn{2}{c|}{\textbf{\fontsize{9.5}{13}\selectfont GoalStep-QnF}} &
\multicolumn{2}{c}{\textbf{\fontsize{9.5}{13}\selectfont HD-EPIC-QnF}} \\[2pt]
 & \multicolumn{2}{c|}{\textbf{IoU = 0.3}}
 & \multicolumn{2}{c}{\textbf{IoU = 0.3}} \\[2pt]
\textbf{\fontsize{9.5}{13}\selectfont Method} & \textbf{R1} & \textbf{R5} & \textbf{R1} & \textbf{R5} \\
\midrule
{OSGNet} & 14.5\greentext{\textsuperscript{+0.2}} & 36.7\greentext{\textsuperscript{+0.7}} & 5.3\greentext{\textsuperscript{+0.4}} & 16.9\greentext{\textsuperscript{+0.7}} \\[2pt]
{ReFocus(OSGNet)} & \textbf{17.9\greentext{\textsuperscript{+3.6}}} & \textbf{42.0\greentext{\textsuperscript{+6.8}}} & \textbf{6.7\greentext{\textsuperscript{+2.2}}} & \textbf{18.6\greentext{\textsuperscript{+4.5}}} \\[2pt]
\midrule
GroundNLQ & 17.7\greentext{\textsuperscript{+0.5}} & 42.2\greentext{\textsuperscript{+1.6}} & 6.6\greentext{\textsuperscript{+0.2}} & 21.3\greentext{\textsuperscript{+0.3}} \\[2pt]
ReFocus(GroundNLQ) & \textbf{20.7\greentext{\textsuperscript{+3.7}}} & \textbf{45.3\greentext{\textsuperscript{+5.0}}} & \textbf{8.2\greentext{\textsuperscript{+1.6}}} & \textbf{25.1\greentext{\textsuperscript{+4.2}}} \\
\bottomrule
\end{tabular}
\caption{Zero-Shot evaluation across feedback datasets when models trained on Ego4D-QnF only.}
\label{tab:zero_shot_results}
\vspace{-0.2cm}

\end{table}

\vspace{-0.4cm}
\paragraph{Zero-Shot Cross Evaluation.} 
Table~\ref{tab:zero_shot_results} shows the performance on GoalStep-QnF and HD-Epic-QnF when the models trained only on Ego4D-QnF, \emph{i.e.}, in the zero-shot setting.
While competing methods show only marginal improvements in this setting, our approach generalizes much better and does not overfit to a specific type of feedback. 
Since GoalStep and HD-EPIC differ from Ego4D-NLQ in video content, they also likely involve different styles of feedback. 
Even so, our approach achieves larger $\Delta$ values on both GoalStep-QnF and HD-EPIC-QnF.

\begin{table}[t]
\centering
\small
\setlength{\tabcolsep}{3.2pt} 
\begin{tabular}{l|cc|cc}
\toprule
 &
\multicolumn{2}{c|}{\textbf{\fontsize{9.5}{13}\selectfont Ego4D-QnF}} &
\multicolumn{2}{c}{\textbf{\fontsize{9.5}{13}\selectfont GoalStep-QnF}} \\[2pt]
 & \multicolumn{2}{c|}{\textbf{IoU = 0.3}}
 & \multicolumn{2}{c}{\textbf{IoU = 0.3}} \\[2pt]
\textbf{\fontsize{9.5}{13}\selectfont Method} & \textbf{R1} & \textbf{R5} & \textbf{R1} & \textbf{R5} \\
\midrule
Gemini-2.5-Flash & 15.7\greentext{\textsuperscript{+1.7}} & 28.7\greentext{\textsuperscript{+0.7}} & 8.7\greentext{\textsuperscript{+2.7}} & 16.0\greentext{\textsuperscript{+1.0}} \\[2pt]
\midrule
ReFocus(OSGNet) & 24.0\greentext{\textsuperscript{+3.0}} & 46.7\greentext{\textsuperscript{+2.7}} & 21.7\greentext{\textsuperscript{+2.7}} & 53.7\greentext{\textsuperscript{+3.7}} \\[2pt]
{ReFocus(GroundNLQ)} & 8.7\greentext{\textsuperscript{+8.7}} & 48.0\greentext{\textsuperscript{+48.0}} & 9.7\greentext{\textsuperscript{+9.7}} & 54.7\greentext{\textsuperscript{+54.7}} \\
\bottomrule
\end{tabular}
\vspace{-0.1cm}
\caption{Performance comparison between Gemini-2.5-Flash and ReFocus models on a small 100 NLQ subset where ReFocus(GroundNLQ) fails with query-only but improves with feedback. Deltas ($\Delta$) of feedback vs query-only performance are shown as superscripts.}
\label{tab:gemini_results}
\vspace{-0.4cm}
\end{table}

\vspace{-0.3cm}
\paragraph{Comparison with Commercial LVLMs.} 
We compare our approach with a commercial LVLM, Gemini-2.5-Flash, on a subset of the test sets. 
We sample 100 NLQs from each of the three QnF datasets where \modelname(GroundNLQ) fails when given only the query, but where at least one user feedback example in the test set helps the model identify the correct response span. 
We then sample three feedback samples for each of the 100 queries, favoring query-relevant spans over irrelevant ones, resulting in 900 QnF samples across the three datasets. 
Table~\ref{tab:gemini_results} shows that, despite the strong R1 performance of Gemini-2.5-Flash, it is unable to significantly improve its predictions when given feedback and a reference span. 
This result suggests that even commercial LVLMs, despite being trained on large-scale multimodal data, still struggle to reason effectively over feedback in long-form videos to produce better predictions. See Supp for more details.

\begin{table}[t]
\centering
\small

\setlength{\tabcolsep}{4pt}
\begin{tabular}{l|cccc}
\toprule
 &
\multicolumn{2}{c}{\textbf{IoU = 0.3}} & 
\multicolumn{2}{c}{\textbf{IoU = 0.5}} \\[2pt]
\textbf{Method} & \textbf{R1} & \textbf{R5} & \textbf{R1} & \textbf{R5} \\
\midrule
GroundNLQ & 29.56 & 56.42 & 21.63 & 43.71 \\
w. FALM & \textbf{33.13} & \textbf{59.70} & \textbf{23.58} & 46.26 \\
w. FALM\textsubscript{\textit{C}} & 31.08 & 57.95 & 22.26 & 44.52 \\
w. FALM\textsubscript{\textit{N}} & 30.89  & 58.03 & 22.38 &  45.02 \\
w. FALM\textsubscript{\textit{T}} & 32.29 & 59.41 & 23.23 & \textbf{46.40} \\
w. FALM w/o Adapter& 32.46 & 58.33 & 23.11 & 45.39 \\
\bottomrule
\end{tabular}
\vspace{-0.15cm}
\caption{
Ablation of our \modelname(GroundNLQ). Evaluated on a subset of Ego4D-QnF containing all types of \searchmodshort labels. %
}

\label{tab:ablation_localization_module}
\end{table}

\subsection{In-depth Model Analysis}
\label{ssec:model_analysis}
\paragraph{Ablations.} 
We present an in-depth ablation study of our approach in Table~\ref{tab:ablation_localization_module} to examine the effect of each component in learning from feedback. 
We start with GroundNLQ and evaluate different components of \searchmodshort. 
First, we study the contribution of each supervision signal used to train FALM (Sec.~\ref{ssec:pnp-module}). We train variants of FALM using a single supervision type: contains, not-contains, and temporal clauses, denoted as FALM\textsubscript{C}, FALM\textsubscript{N}, and FALM\textsubscript{T}, respectively. 
We observe that each clause type improves the model, with the temporal clause being the most helpful. 
Combining all of them leads to further gains, since different feedback may contain different types of information. 
We also evaluate whether integrating \searchmodshort with the proposed EM adapter is beneficial. The results (last row in Table~\ref{tab:ablation_localization_module}) show a drop in performance when the adapter is removed.

\begin{figure}[t]
    \centering
    
    \begin{subfigure}[t]{0.75\columnwidth}
        \centering
        \includegraphics[width=\linewidth]{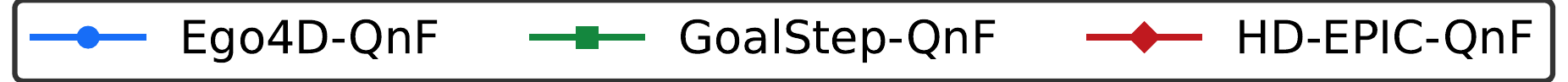} %
        \label{fig:legend}
    \end{subfigure}%
    \vspace{-0.35cm}

    \begin{subfigure}[t]{0.49\columnwidth}
        \centering
        \includegraphics[width=\linewidth]{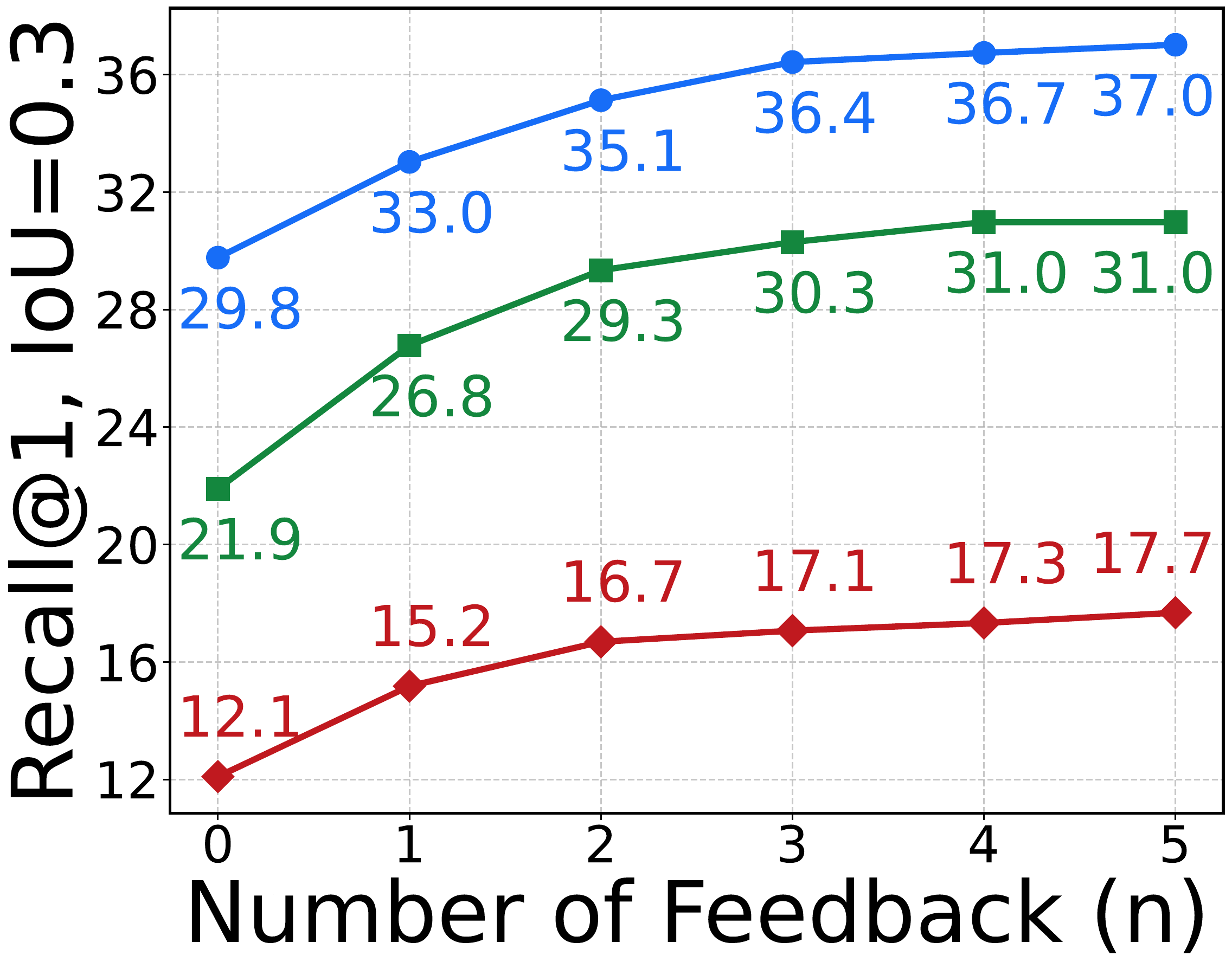} 
        \label{fig:a_r1}
    \end{subfigure}%
    \hfill
    \begin{subfigure}[t]{0.49\columnwidth}
        \centering
        \includegraphics[width=\linewidth]{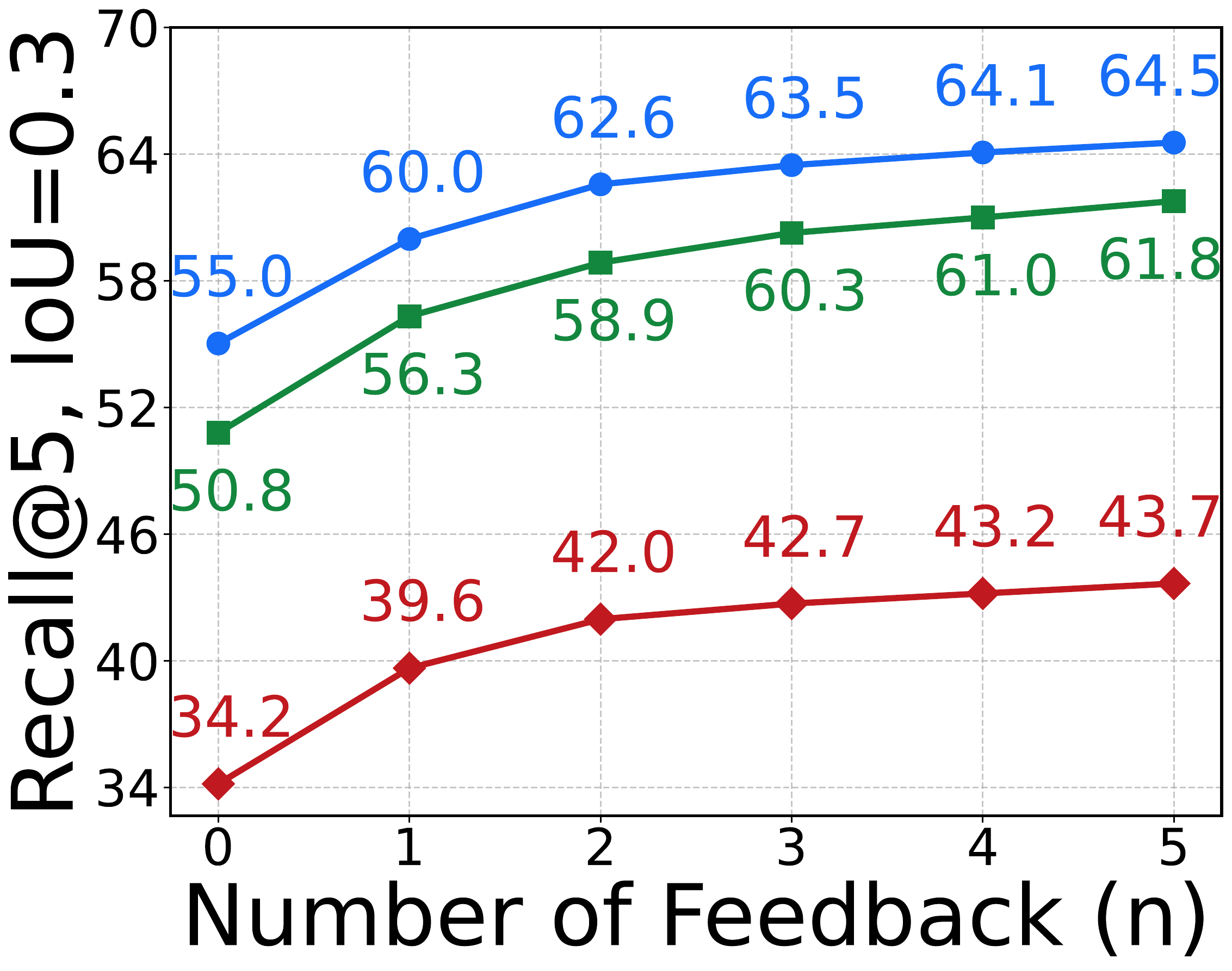} 
        \label{fig:a_r5}
    \end{subfigure}
     \vspace{-0.35cm}

    \caption{
        Multi-Turn Feedback evaluation of our \modelname(GroundNLQ) across the three datasets.
    }
    \label{fig:multi_feedback_eval}
\end{figure} 

\begin{figure*}[th!]
    \centering
    \includegraphics[width=0.99
    \linewidth]{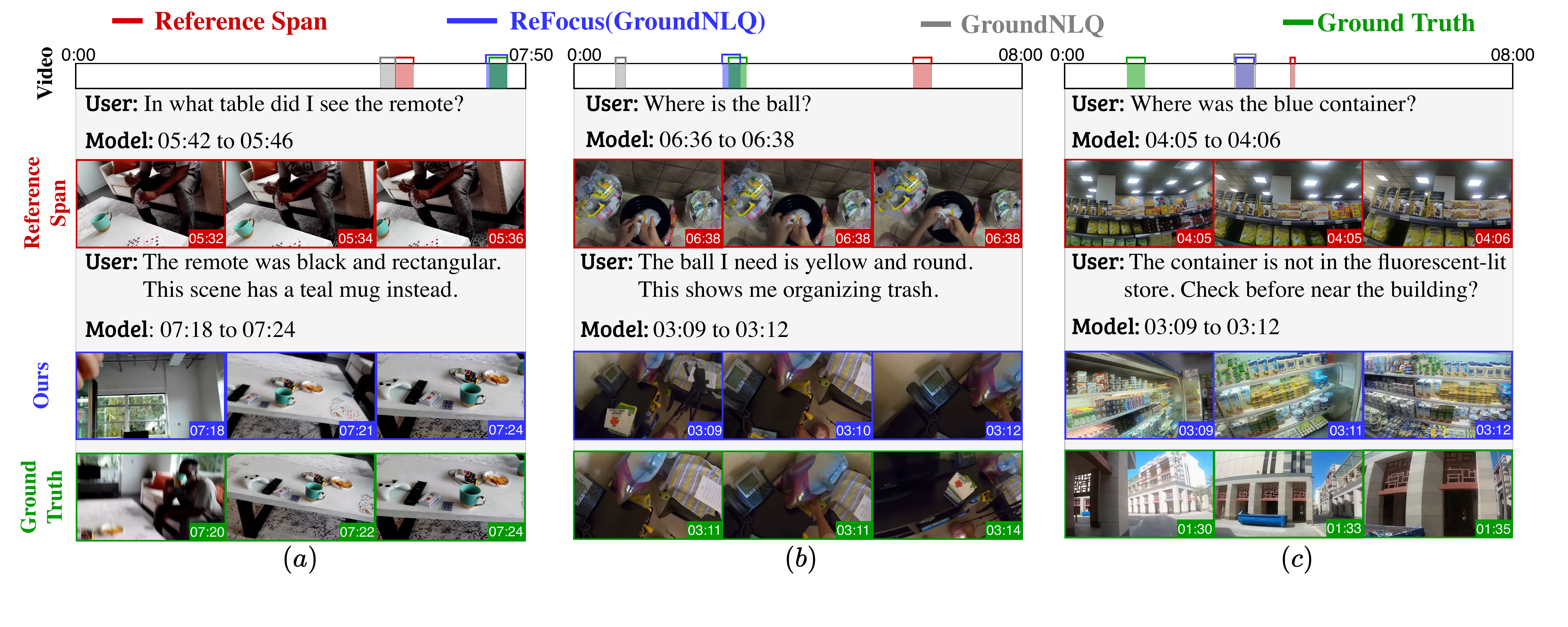}
    \vspace{-0.5cm}
    \caption{
        Qualitative results for GroundNLQ and our ReFocus(GroundNLQ) when given feedback. Examples (a) and (b) show cases where \modelname(GroundNLQ) improves with feedback, whereas (c) shows a failure case.
    }
    \label{fig:qualitative_res}
    \vspace{-0.3cm}
\end{figure*}

\vspace{-0.2cm}
\paragraph{Multi-Turn Feedback.} 
While we have focused so far on the single-turn case, we evaluate here the ability of \modelname~to handle multi-turn feedback in a zero-shot manner, following the proposed extension in Sec.~\ref{ssec:pnp-module}. 
In our evaluation, we average performance over five different random samplings of $n$ feedback per query. 
Figure~\ref{fig:multi_feedback_eval} shows the performance of \modelname(GroundNLQ) across the three datasets with multiple feedback. 
Interestingly, our approach improves as the number of feedback turns increases on all datasets, even though it was not trained in this setting. 
We observe substantial gains up to the third or fourth feedback turn, after which performance plateaus. %

\begin{table}[t!]
\centering
\small
\setlength{\tabcolsep}{4pt}
\begin{tabular}{l|cccc}
    \toprule
    &
    \multicolumn{2}{c}{\textbf{IoU = 0.3}} & 
    \multicolumn{2}{c}{\textbf{IoU = 0.5}} \\[2pt]
    \textbf{Feedback Type} & $\Delta$\textbf{R1} & $\Delta$\textbf{R5} & $\Delta$\textbf{R1} & $\Delta$\textbf{R5} \\
    \midrule
    Generated & 8.6   & 50.0  & 5.4 & 31.0  \\
    {Human}         & 5.8   & 34.4  & 3.4 & 20.4 \\
    \bottomrule
\end{tabular}
\caption{
    {Comparison of our ReFocus(GroundNLQ) on generated vs human feedback on examples from Ego4d-QnF where the method fails with query-only. 
    The reported metrics are absolute improvement in Recall when using feedback vs without.} 
}
\vspace{-0.2cm}
\label{tab:human_feedback_eval}
\end{table}

\paragraph{Comparison with Human Feedback.}\label{ssec:human-eval} 
We collected feedback from human users to compare it with the quality of our generated feedback. 
From Ego4D-QnF, we sample unique NLQ and reference span pairs where \modelname(GroundNLQ) fails but improves with LVLM-generated feedback. 
We ask 11 users to assume the role of the person wearing the camera in the egocentric video and provide feedback for the reference spans as if they were trying to recall the correct response span. 
Users are instructed not to answer the queries directly, but instead to guide the model toward the correct response span they are trying to recall. 
In total, we collect 500 unique user feedback for 271 NLQ and reference span pairs. 
Table~\ref{tab:human_feedback_eval} shows the evaluation of our approach using generated and human feedback. 
We observe that our approach can leverage human feedback and improve its predictions, which suggests that our proposed feedback generation recipe is effective at producing realistic and helpful feedback that generalizes to human-style inputs.  
Additionally, we see there is still a gap in absolute improvement between generated and human feedback, indicating room for further improvement. Methods trained with human feedback may be able to recover additional gains.

\vspace{-0.3cm}
\paragraph{Qualitative Results and Failures.} 
Figure~\ref{fig:qualitative_res} shows qualitative examples from the Ego4D-QnF dataset.
Examples (a) and (b) show that \modelname(GroundNLQ) improves over GroundNLQ when given additional user feedback about object attributes in contrast to the reference span. 
Example (c) shows an interesting failure case for our approach. 
The user feedback suggests that the blue container is not inside the store, but near a building, and also indicates that the model should search before the reference span. 
While the model correctly shifts its attention to an earlier moment, it is confused by the many blue food containers on the shelves and fails to infer that those shelves are inside the same store.

\section{Conclusion}
This work addresses an unexplored aspect of episodic memory search with natural language queries: the interactive nature of the task. User questions can be incomplete or ambiguous, and model predictions can be incorrect. In such cases, the model should be able to incorporate user feedback to improve its predictions. 
We introduce the task of interactive episodic memory with user feedback (EM-QnF), a suitable recipe for user feedback generation without manual annotations, and an effective feedback alignment module (FALM) that can be integrated into different EM-NLQ models, leading to significant improvements across multiple challenging benchmarks.

\vspace{-0.3cm}
\paragraph{Acknowledgements}
We thank Santhosh Ramakrishnan for valuable feedback during the early stages of the project. This work was partially supported by NSF 2421782 and MPS-AI-00010515.

\clearpage

{
    \small
    \bibliographystyle{ieeenat_fullname}
    \bibliography{main}
}

\clearpage
\setcounter{page}{1}

\section{Supplementary Material}
\label{sec:supplementary}
In this supplementary material, we provide further information about dataset generation, model evaluation, and limitations of our work. 
All code and datasets can be found on our project page: \url{https://nsubedi11.github.io/refocus}.

The supplementary material is sectioned as follows: 
\begin{itemize}
\item Sec \ref{ssec:supp_video}: Supplementary Video with qualitative results.
\item  Sec \ref{ssec:supp_additional_model_analysis}: Additional Model Analysis.
\item Sec \ref{ssec:supp_commercial_lvlm}: Commercial LVLM Evaluation. 
\item Sec \ref{ssec:supp_limitations_future_work}: Limitations and Future Work.
\item Sec \ref{ssec:supp_human_user_feedback_collection}: Human User Feedback Collection.

\item Sec \ref{ssec:supp_nlq_generation}: Query generation for GoalStep and HD-EPIC.
\item Sec \ref{ssec:supp_qnf_datasets}: Feedback generation for QnF Datasets.
\item Sec \ref{ssec:supp_eval_setup}: Additional Evaluation Setup Details.

\end{itemize}

\subsection{Supplementary Video}
\label{ssec:supp_video}
We provide additional qualitative results and comparisons for our approach \modelname\ in video format on our project page: \url{https://nsubedi11.github.io/refocus}. 
The page showcases a variety of scenarios, highlights interactions with user feedback, and includes comparisons with baseline methods. We also present failure cases where our approach struggles to effectively utilize the information provided through user feedback.

\subsection{Additional Model Analysis}
\label{ssec:supp_additional_model_analysis}

\paragraph{Performance of User Feedback on Incorrect First Predictions.}
Here, we report the performance of each method on the subset of the QnF dataset containing only the queries that the method fails to solve when given the query alone, i.e., those for which the method achieves Recall@5 at IoU=0.3 of 0.
Since each model is evaluated on its own subset, this analysis measures how effectively it leverages user feedback.
Table~\ref{tab:supp_delta_results} reports the performance of the GroundNLQ method on these subsets.
Our approach \modelname~leads to significant improvements across all datasets and metrics, demonstrating that it effectively leverages user feedback to correct the model's initial mistakes when using only the query.

\begin{table*}[ht]
\centering
\small
\setlength{\tabcolsep}{3.5pt} 
\begin{tabular}{l|ccccc|ccccc|ccccc}
\toprule
 &
\multicolumn{5}{c|}{\textbf{\fontsize{9.5}{13}\selectfont Ego4D-QnF}} & 
\multicolumn{5}{c|}{\textbf{\fontsize{9.5}{13}\selectfont GoalStep-QnF}} &
\multicolumn{5}{c}{\textbf{\fontsize{9.5}{13}\selectfont HD-EPIC-QnF}} \\[2pt]
 & &  \multicolumn{2}{c}{\textbf{IoU = 0.3}} & \multicolumn{2}{c|}{\textbf{IoU = 0.5}}
 & & \multicolumn{2}{c}{\textbf{IoU = 0.3}} & \multicolumn{2}{c|}{\textbf{IoU = 0.5}}
 & & \multicolumn{2}{c}{\textbf{IoU = 0.3}} & \multicolumn{2}{c}{\textbf{IoU = 0.5}} \\[2pt]
\textbf{\fontsize{9.5}{13}\selectfont Method} &
\textbf{\#$\mathcal{F}$}&$\Delta$\textbf{R1} & $\Delta$\textbf{R5} & $\Delta$\textbf{R1} & $\Delta$\textbf{R5} & \textbf{\#$\mathcal{F}$} & $\Delta$\textbf{R1} & $\Delta$\textbf{R5} & $\Delta$\textbf{R1} & $\Delta$\textbf{R5} & \textbf{\#$\mathcal{F}$} &  $\Delta$\textbf{R1} & $\Delta$\textbf{R5} & $\Delta$\textbf{R1} & $\Delta$\textbf{R5} \\
\midrule
GroundNLQ &  9898 & 0.42 & 6.59 & 0.18 & 3.22 &  6620 & 0.44 & 6.15 & 0.17 & 3.94 &  10069 & 0.22 & 4.53 & 0.06 & 2.11 \\[2pt]

ReFocus(GroundNLQ) & 9863 & \textbf{2.72} & \textbf{16.42} & \textbf{1.40} & \textbf{9.64} & 6940 & \textbf{3.17} & \textbf{18.87} & \textbf{1.93} & \textbf{13.46} &  9874 & \textbf{1.51} & \textbf{12.70} & \textbf{0.65} & \textbf{6.52} \\

\bottomrule
\end{tabular}
\caption{Model performance delta relative to its own failure cases with query only. 
Each method is test with user feedback for NLQs where the method fails (achieves R5, IoU=0.3 = 0) when only the NLQ is given.}
\label{tab:supp_delta_results}
\end{table*}

\paragraph{Feedback on the Model's Own Failure Spans.}
We evaluate whether our reference span sampling strategy generalizes to model errors by constructing a targeted setting in which feedback is generated only for queries that the model initially fails and using that wrong prediction as the reference span.
Specifically, for ReFocus(GroundNLQ), we identify queries whose top-1 prediction under the query-only setting has IoU$<0.3$, and generate user feedback using the recipe described in Sec.~\ref{ssec:recipe-data}, treating the model's top-1 prediction as the reference span.
Table~\ref{tab:supp_model_failure_reference_delta_results} shows the results.
Across all datasets, we observe substantial positive gains in both R1 and R5 at IoU thresholds of 0.3 and 0.5.
Notably, incorporating feedback for these spans leads to larger gains (e.g., +12.77 R1@0.3 on Ego4D-NLQ and +13.57 R1@0.3 on GoalStep-Q) than those on the EM-QnF evaluation set reported in Table~\ref{tab:combined_feedback_results}.
These results indicate that ReFocus can effectively leverage feedback to correct its own high-confidence errors.

\begin{table}[htbp]
\centering
\small
\setlength{\tabcolsep}{4pt}

\begin{tabular}{l c c c c c}
\toprule
\textbf{Dataset} & \textbf{\#$\mathcal{F}$/$\mathcal{Q}$} 
& \multicolumn{2}{c}{\textbf{IoU = 0.3}} 
& \multicolumn{2}{c}{\textbf{IoU = 0.5}} \\
& & $\Delta$R1 & $\Delta$R5 & $\Delta$R1 & $\Delta$R5 \\
\midrule

Ego4D-NLQ  & 12620/3196 & 12.77 & 43.7  & 8.11  & 30.14 \\
GoalStep-Q & 9355/2341  & 13.57 & 44.55 & 10.04 & 34.86 \\
HD-EPIC-Q  & 10319/2634 & 7.61  & 31.99 & 4.24  & 19.61 \\

\bottomrule
\end{tabular}

\caption{Performance delta for \textbf{ReFocus (GroundNLQ)} with feedback generated from its own failure spans, relative to using query only. User feedback is generated using the recipe in \ref{ssec:recipe-data}, where feedback is constructed from top-1 predictions for queries where the model fails (R1@IoU=0.3 = 0). \#$\mathcal{F}$/$\mathcal{Q}$ denotes number of feedback instances and number of incorrect queries.}
\label{tab:supp_model_failure_reference_delta_results}
\end{table}

\begin{table}[htbp]
\centering

\small
\begin{tabular}{l|cccc}
\toprule
 & 
\multicolumn{2}{c}{\textbf{IoU = 0.3}} & 
\multicolumn{2}{c}{\textbf{IoU = 0.5}} \\[2pt]
\textbf{Method} & \textbf{R1} & \textbf{R5} & \textbf{R1} & \textbf{R5} \\
\midrule
\multicolumn{5}{l}{\textit{Query-Irrelevant Reference Span}} \\
GroundNLQ & 30.70 & 57.39 & 22.21 & 44.22 \\
\modelname(GroundNLQ) & 34.09 & 61.03 & 24.42 & 47.32 \\
$\Delta$ & \greentext{+3.39} & \greentext{+3.64} & \greentext{+2.21} & \greentext{+3.10} \\
\midrule
\multicolumn{5}{l}{\textit{Query-Relevant Reference Span}} \\
GroundNLQ & 18.29 & 40.77 & 12.44 & 30.00 \\
\modelname(GroundNLQ) & 22.89 & 45.32 & 15.89 & 33.82 \\
$\Delta$ &  \greentext{+4.60} &  \greentext{+4.55} &  \greentext{+3.45} &  \greentext{+3.82} \\
\bottomrule
\end{tabular}
\caption{Impact of user feedback on NLQ with different reference span relevance on Ego4D-QnF dataset.}

\label{tab:supp_feedback_relevance}
\end{table}

\begin{table*}[pbht]
\centering
\small
\setlength{\tabcolsep}{3pt}

\begin{tabular}{@{}l|
    P{1.1cm}P{1cm}P{1cm}P{1cm}P{1cm}P{1.3cm}P{1.1cm}P{1.0cm}P{1.0cm}|
    P{1.3cm}@{}}
\toprule

& \multicolumn{9}{c|}{Object / place queries}
& \multicolumn{1}{c}{People queries} \\
\cmidrule(lr){2-10} \cmidrule(lr){11-11}

{Method}
& {Where is X before/ after Y?}
& { Where did I put X?}
& {Where is X?}
& {What did I put in X?}
& { How many X's?}
& {In what location did I see X?}
& {What X did I Y?}
& {What X is Y?}
& {State?}
& {Who did I interact with during Y?}
\\
\midrule						
OSGNet &
20.5\redtext{\textsuperscript{-0.3}} &
18.1\greentext{\textsuperscript{+1.0}} &
15.8\redtext{\textsuperscript{-0.0}} &
22.7\greentext{\textsuperscript{+0.7}} &
26.4\greentext{\textsuperscript{+0.2}} &
15.2\greentext{\textsuperscript{+1.0}} &
28.3\greentext{\textsuperscript{+1.1}} &
14.5\greentext{\textsuperscript{+0.6}} &
28.5\redtext{\textsuperscript{-0.8}} &
18.3\greentext{\textsuperscript{+0.9}} \\[2pt]
					
ReFocus(OSGNet) &
21.2\redtext{\textsuperscript{-0.0}} &
19.5\greentext{\textsuperscript{+1.0}} &
16.7\greentext{\textsuperscript{+2.1}} &
25.4\greentext{\textsuperscript{+3.2}} &
31.8\greentext{\textsuperscript{+3.6}} &
15.8\greentext{\textsuperscript{+1.8}} &
27.7\greentext{\textsuperscript{+3.4}} &
19.1\greentext{\textsuperscript{+1.9}} &
31.6\greentext{\textsuperscript{+1.8}} &
22.6\redtext{\textsuperscript{-0.0}} \\[2pt]
\midrule
GroundNLQ &
23.9\redtext{\textsuperscript{-0.9}} &
21.9\greentext{\textsuperscript{+0.3}} &
10.7\greentext{\textsuperscript{+0.1}} &
23.8\redtext{\textsuperscript{-0.2}} &
28.6\redtext{\textsuperscript{-0.4}} &
9.6\greentext{\textsuperscript{+0.4}} &
30.7\greentext{\textsuperscript{+1.0}} &
16.3\greentext{\textsuperscript{+0.1}} &
29.6\greentext{\textsuperscript{+1.6}} &
{19.8}\greentext{\textsuperscript{+0.7}} \\[2pt]

ReFocus(GroundNLQ) &
{25.5}\greentext{\textsuperscript{+0.6}} &
{25.2}\greentext{\textsuperscript{+3.3}} &
12.2\greentext{\textsuperscript{+1.1}} &
{27.8}\greentext{\textsuperscript{+4.7}} &
{31.4}\greentext{\textsuperscript{+2.4}} &
10.3\greentext{\textsuperscript{+0.9}} &
{34.3}\greentext{\textsuperscript{+3.4}} &
{19.3}\greentext{\textsuperscript{+1.5}} &
{33.4}\greentext{\textsuperscript{+3.6}} &
19.0\redtext{\textsuperscript{-1.0}} \\[2pt]
\midrule
\end{tabular}

\caption{Evaluation with user feedback on different query types in Ego4D-QnF. Deltas ($\Delta$) of feedback vs query-only performance are shown as superscripts.}
\label{tab:supp_query_types}
\end{table*}

\begin{table}[thbp]
\centering
\small

\setlength{\tabcolsep}{4pt}
\begin{tabular}{l|cccc}
\toprule
 &
\multicolumn{2}{c}{\textbf{IoU = 0.3}} & 
\multicolumn{2}{c}{\textbf{IoU = 0.5}} \\[2pt]
\textbf{Method} & \textbf{R1} & \textbf{R5} & \textbf{R1} & \textbf{R5} \\
\midrule
GroundNLQ & 29.56 & 56.42 & 21.63 & 43.71 \\
\midrule
w. \searchmodshort & 33.13 & 59.70 & 23.58 & 46.26 \\
w. \searchmodshort\textsubscript{\textit{C}} & 31.08 & 57.95 & 22.26 & 44.52 \\
w. \searchmodshort\textsubscript{\textit{N}} & 30.89  & 58.03 & 22.38 &  45.02 \\
w. \searchmodshort\textsubscript{\textit{T}} & 32.29 & 59.41 & 23.23 & 46.40 \\
\midrule
w. {GT} \searchmodshort  & \textbf{40.93}  & \textbf{68.92}  & \textbf{30.55}  & \textbf{55.73}  \\
w. {GT} \searchmodshort \textsubscript{\textit{C}}  & 35.88  & 63.75  & 26.18 & 50.52 \\
w. {GT} \searchmodshort \textsubscript{\textit{N}}  & 34.62 & 63.09  & 24.99 & 49.57  \\
w. {GT} \searchmodshort \textsubscript{\textit{T}}  & 32.19  & 59.12  & 23.29   & 46.06   \\
\bottomrule
\end{tabular}
\caption{
Additional ablation on \modelname(GroundNLQ). 
GT \searchmodshort denotes using ground truth \searchmodshort labels instead of predicted ones.
Evaluation results are on a subset of Ego4D-QnF containing all types of \searchmodshort labels (see Sec \ref{ssec:pnp-module}).
}
\label{tab:supp_ablation_localization_module}
\vspace{-0.2cm}
\end{table}

\begin{table}[ht]
\centering
\setlength{\tabcolsep}{3.2pt}
\small
\begin{tabular}{lcccc}
\hline
Method & Params(B)$\downarrow$  & TFLOPs$\downarrow$  & Inf. Speed$\uparrow$  \\
\hline
TimeChat \cite{timechat} & 7.97 &  238  & 0.45  \\
UniTime \cite{unitime} & 8.29 &  142 & 0.07 \\
OSGNet \cite{osgnet} & 0.12 & 0.58 & 5.36 \\
ReFocus(OSGNet) & 0.14 & 0.61 & 4.92 \\
GroundNLQ \cite{groundnlq}& 0.05 & 1.24 & 29.12  \\
ReFocus(GroundNLQ) & 0.08 & 1.64 & 23.52 \\
\hline
\end{tabular}
\caption{Efficiency comparison between different methods. Inf. Speed represents average inference speed in number of query (with/without user feedback) per seconds. We assume video features are pre-extracted for TimeChat and UniTime and both video and text features are pre-extracted for EM-NLQ methods.}
\label{tab:efficiency_comparison}
\end{table}

\begin{figure}[t]
    \centering
    
    \begin{subfigure}[t]{0.75\columnwidth}
        \centering
        \includegraphics[width=\linewidth]{figs/recall_legend_datasets.pdf} %
        \label{fig:supp_legend}
    \end{subfigure}%
    \vspace{-0.35cm}

    \begin{subfigure}[t]{0.49\columnwidth}
        \centering
        \includegraphics[width=\linewidth]{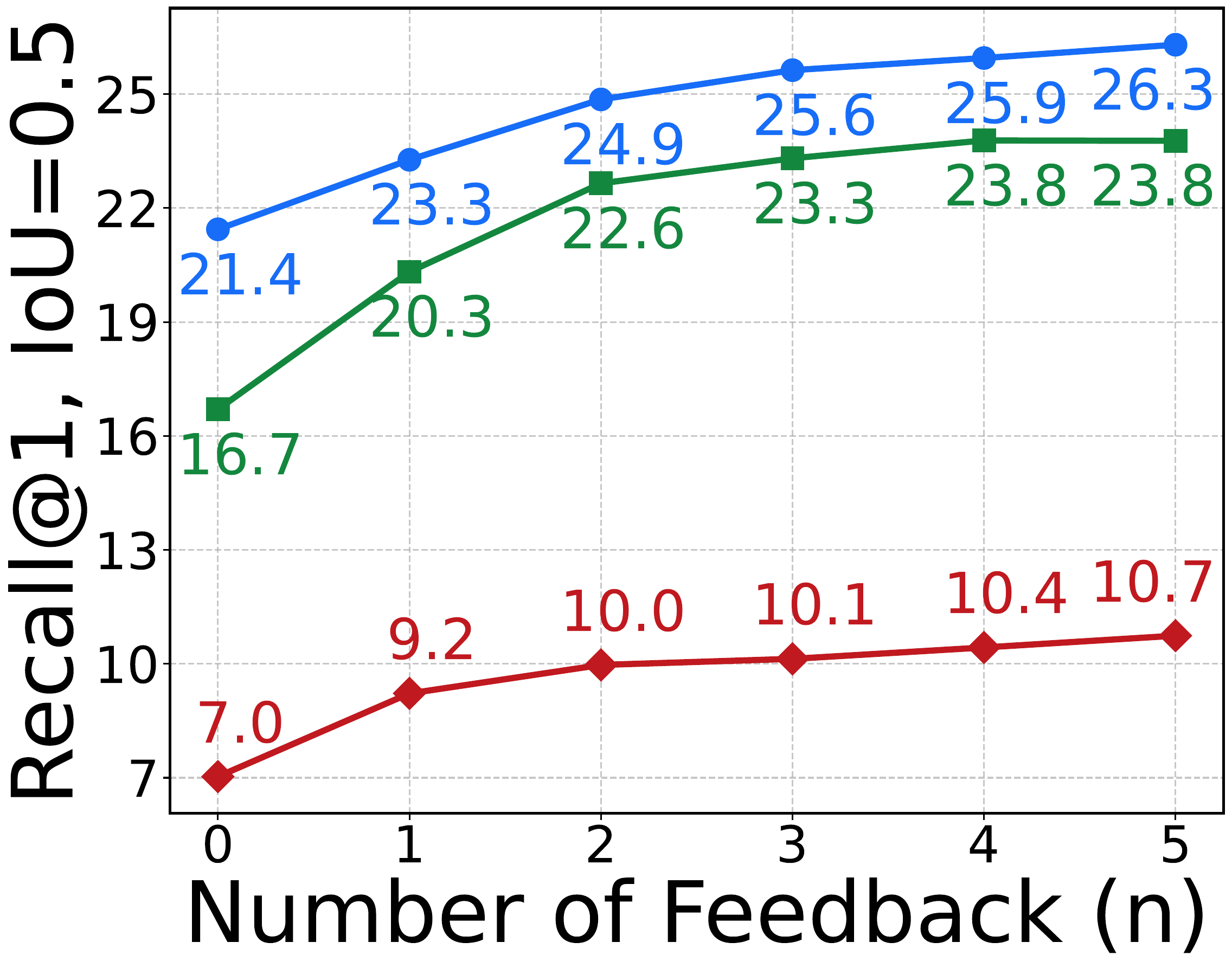} 
        \label{fig:supp_a_r1}
    \end{subfigure}%
    \hfill
    \begin{subfigure}[t]{0.49\columnwidth}
        \centering
        \includegraphics[width=\linewidth]{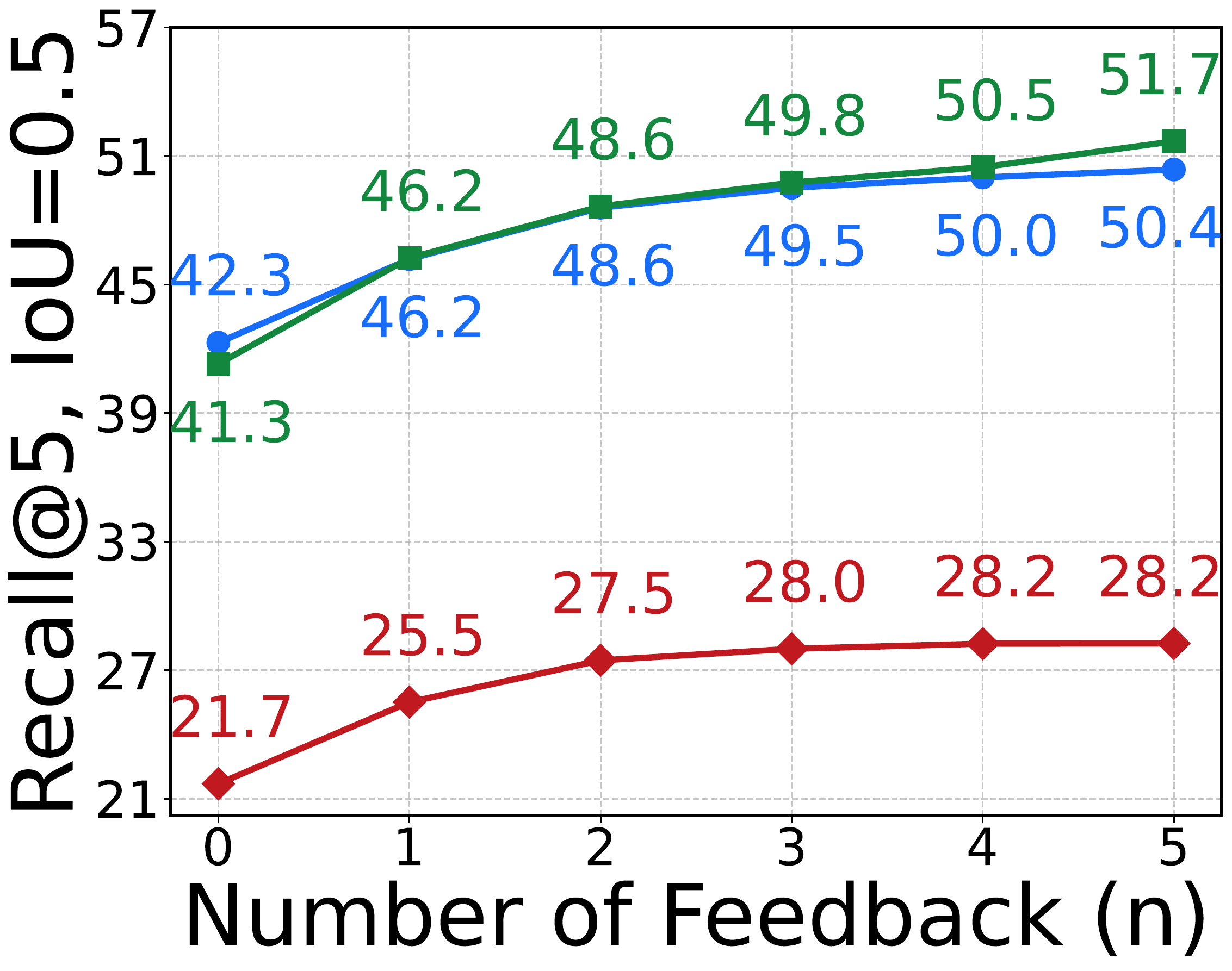} 
        \label{fig:supp_a_r5}
    \end{subfigure}

    \caption{
        Multi-Turn Feedback evaluation of our \modelname(GroundNLQ) across the three datasets at IoU=0.5.
    }
    \label{fig:supp_multi_feedback_eval}
    
\end{figure}

\begin{figure}[t]
    \centering
    
    \begin{subfigure}[t]{0.35\columnwidth}
        \centering
        \includegraphics[width=\linewidth]{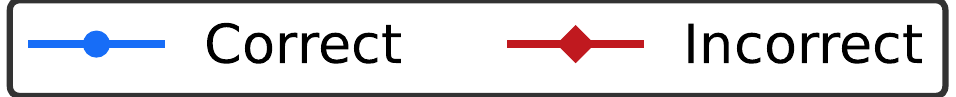} %
    \end{subfigure}%

    \begin{subfigure}[t]{0.49\columnwidth}
        \centering
        \includegraphics[width=\linewidth]{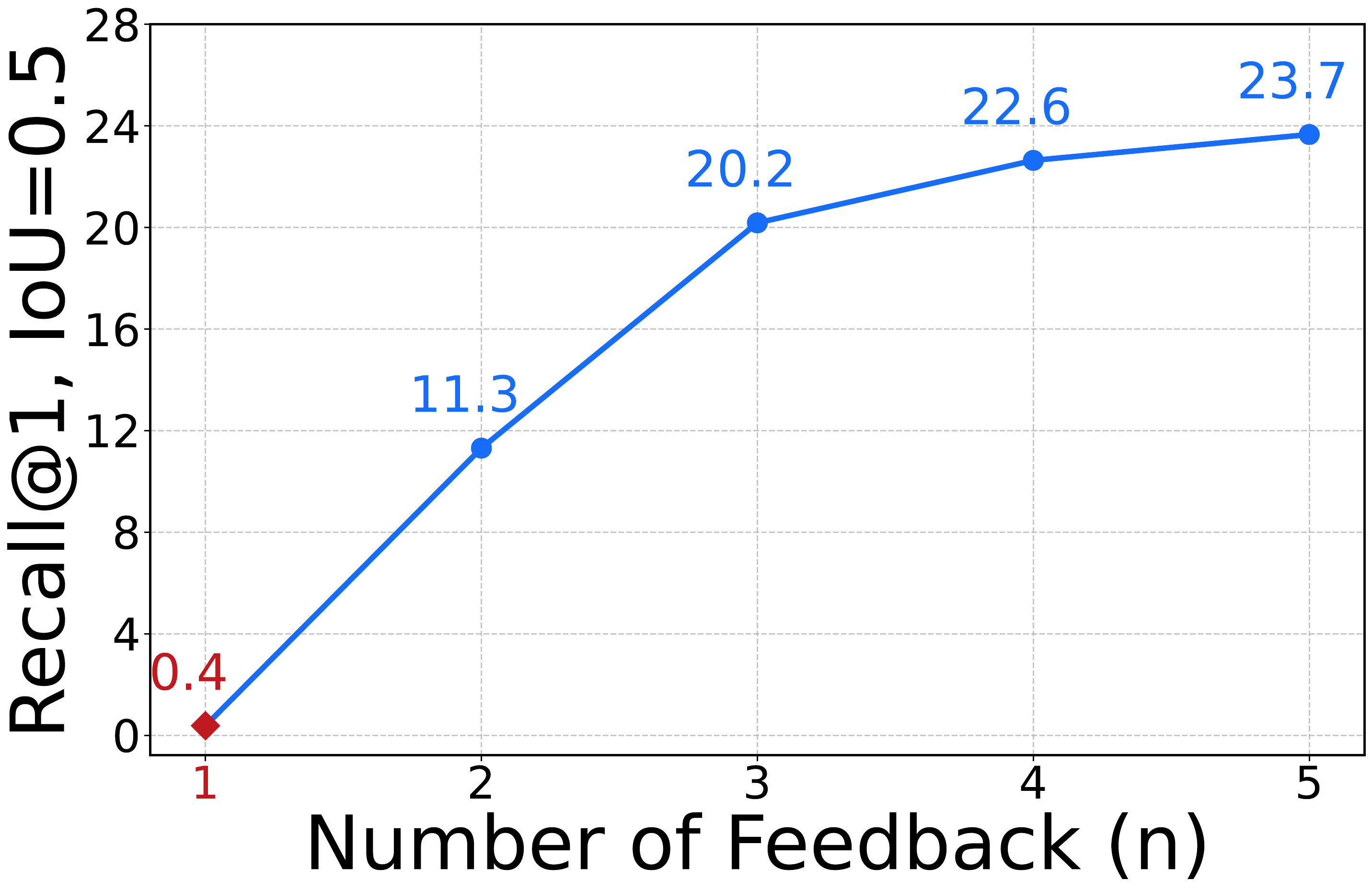} 
        \label{fig:supp_rebuttal_r1}
    \end{subfigure}%
    \hfill
    \begin{subfigure}[t]{0.49\columnwidth}
        \centering
        \includegraphics[width=\linewidth]{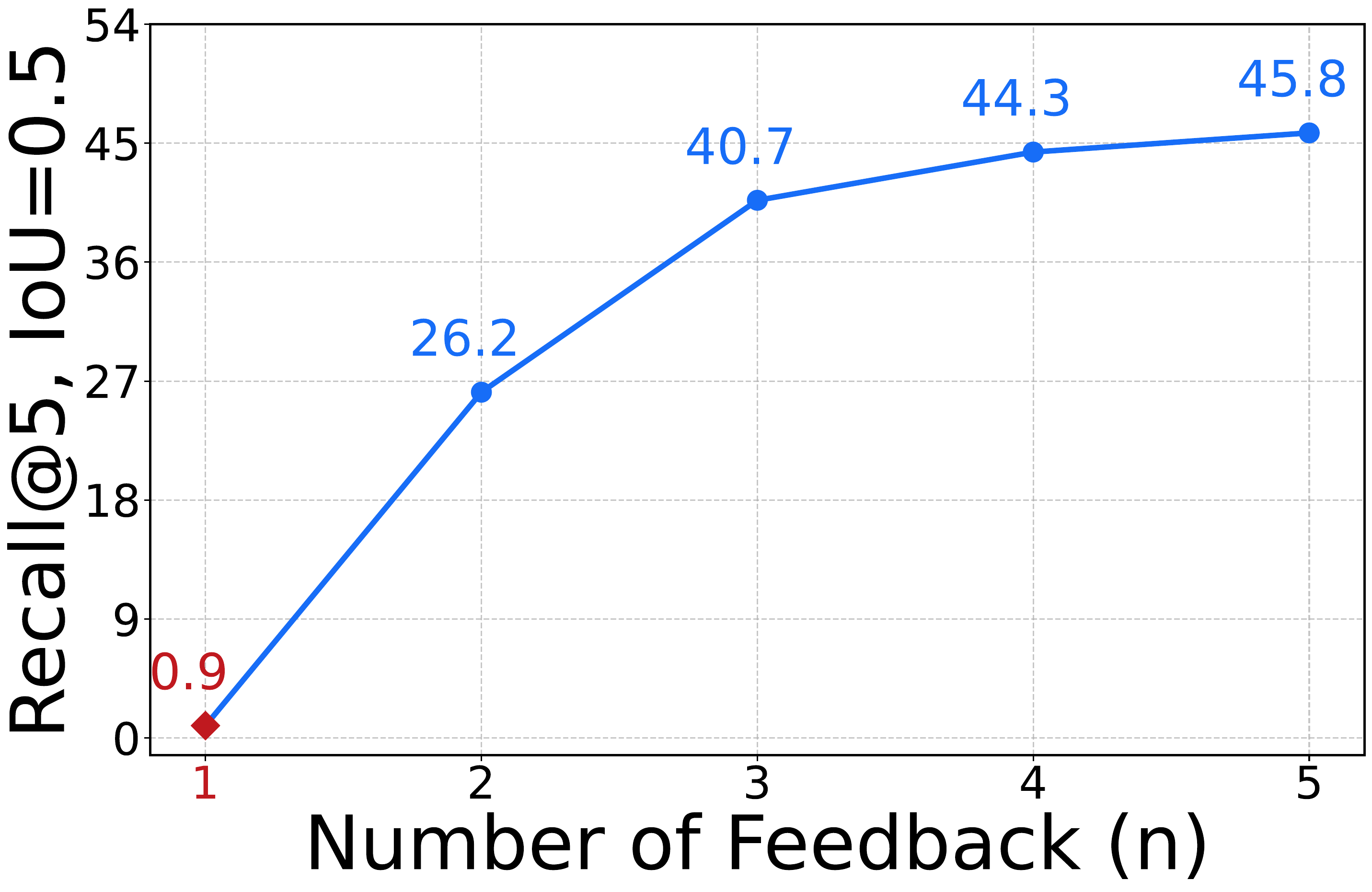} 
        \label{fig:supp_rebuttal_r5}
    \end{subfigure}

    \caption{
    Noisy Multi-Turn Feedback evaluation of our ReFocus(GroundNLQ) on Ego4D-QnF at IoU=0.5.
    Given an initial wrong feedback (red) followed by correct ones (blue), ReFocus(GroundNLQ) is able to recover in multi-turn setting. 
    }
    \label{fig:rebut_noisy_multi_feedback_eval}
\end{figure}

\paragraph{Multi-Turn Feedback Evaluation.}
Figure~\ref{fig:multi_feedback_eval} in the main manuscript shows the multi-turn feedback results for IoU$=0.3$.
Here, we also report the results for IoU$=0.5$.
We report the average recall after 5 different samplings of $n$ feedback instances for each NLQ.
Figure~\ref{fig:supp_multi_feedback_eval} shows the multi-turn feedback evaluation on the three datasets: Ego4D-QnF, GoalStep-QnF, and HD-EPIC-QnF.

We observe the same trend as in Figure~\ref{fig:multi_feedback_eval}: the first few rounds of feedback improve recall substantially, and performance then plateaus after the third or fourth feedback instance.
These results further demonstrate that our approach can effectively leverage multiple rounds of feedback, even without training on sequential multi-feedback data.

\paragraph{Effect of Noise in Multi-Turn Feedback.}
We also study the effect of noisy feedback and evaluate whether our method can recover from it.
To do so, we first provide incorrect temporal feedback, i.e., feedback that points the model in the wrong direction, and then follow it with correct feedback.
Figure~\ref{fig:rebut_noisy_multi_feedback_eval} shows the ability of ReFocus(GroundNLQ) to recover from incorrect feedback on Ego4D-QnF.
As expected, after the initial incorrect feedback, the model achieves less than 1\% Recall.
However, with subsequent feedback, the model recovers and reaches 22.6\% Recall@1 after the third correct feedback, which is comparable to the non-noisy setting.
Notably, although our model is never trained with incorrect feedback, it can still recover from such noise, demonstrating the robustness of our approach in realistic settings where users may some times provide erroneous feedback.

\paragraph{Impact of Feedback by Reference Span Type.}
We analyze the impact of user feedback generated from different types of reference spans.
As discussed in Sec.~\ref{ssec:recipe-data} and Sec.~\ref{ssec:supp_qnf_datasets}, we generate user feedback by sampling different reference span types.
Query-relevant reference spans are either visually similar to the target response spans or correspond to incorrect top-1 predictions from an EM-NLQ model.
Query-irrelevant reference spans are either response spans from other queries in the same video or randomly sampled spans.

Table~\ref{tab:supp_feedback_relevance} compares the performance of GroundNLQ and our approach, \modelname(GroundNLQ), on Ego4D-QnF for these two reference span types.
Overall, performance is lower on examples with query-relevant reference spans than on those with query-irrelevant spans, since query-relevant spans are sampled from difficult NLQs for which the pretrained EM-NLQ model fails to localize the correct response span.
Notably, our approach improves performance in both settings, and the absolute gain over the EM-NLQ model without \modelname~is larger on the query-relevant subset than on the query-irrelevant subset.
This suggests that our approach helps in both settings and is more effective at leveraging feedback from query-relevant reference spans, which may provide more disambiguating information than query-irrelevant ones.

\paragraph{Impact of Feedback by Query Template.}
Table~\ref{tab:supp_query_types} reports the performance of our method grouped by the query types defined in Ego4D~\cite{ego4d}.
We observe that user feedback yields the largest improvements for queries about objects rather than object locations.
Interestingly, \modelname~keep or slightly degrades performance on queries about people and asking whom the user interacted with.
As in earlier evaluations, \modelname~leads to substantially larger gains than the baselines on most query types.

\paragraph{ReFocus Ablation with Ground-Truth FALM Labels.}
Here, we extend the ablation in Table~\ref{tab:ablation_localization_module} from Sec.~\ref{ssec:model_analysis}.
We consider several settings in which the EM-NLQ model is provided with ground-truth FALM pseudo-labels, used to train FALM, instead of the predicted FALM output $P$, to estimate the upper-bound performance of FALM.
We evaluate pseudo-labels derived from each of the three clause types extracted from the user feedback.
Table~\ref{tab:supp_ablation_localization_module} shows these additional ablations for our proposed approach, \modelname(GroundNLQ).

Compared to the trained FALM, using ground-truth \textit{contains} and \textit{not-contains} labels, $L^c$ and $L^k$, yields larger gains than using ground-truth \textit{temporal} labels.
We also observe that trained and ground-truth \textit{temporal} \searchmodshort\ achieve similar performance.
As in the trained setting, combining signals from all three label types leads to even higher performance.
Comparing FALM with ground-truth FALM indicates that there is still room for improvement in modeling the feedback alignment signal, especially for the \textit{contains} and \textit{not-contains} signals.

\paragraph{Efficiency Comparison.}
Table~\ref{tab:efficiency_comparison} compares the efficiency of the baselines and our approach.
As expected, LVLM-based methods are inefficient, especially because they require multiple autoregressive forward passes to generate new tokens.
Notably, our approach is lightweight, adding only \(\sim\)8\% overhead to the inference speed of OSGNet and \(\sim\)20\% to that of GroundNLQ, and the computation cost and inference speed of our method is still much better than LVLM-based methods.

\begin{figure*}[thbp]
    \centering
    \includegraphics[width=0.99
    \linewidth]{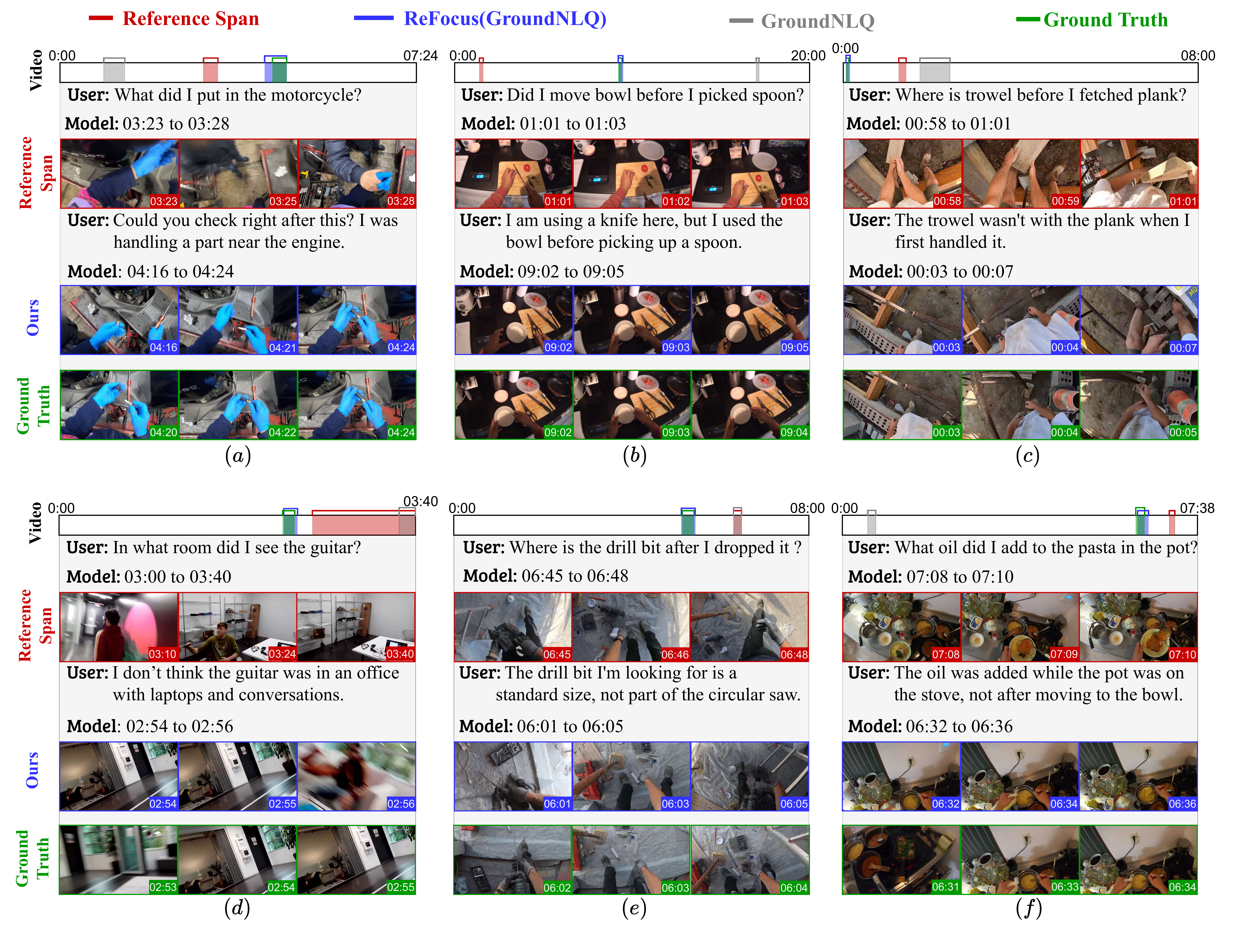}
    \caption{Additional Qualitative Results. These examples show our method improving on the baseline.}
    \label{fig:supp_good_qualitative}
\end{figure*}

\paragraph{Additional Qualitative Results and Failure Cases.}
Figure~\ref{fig:supp_good_qualitative} shows examples in which the baseline GroundNLQ model fails, while our approach, \modelname(GroundNLQ), correctly identifies the response span.
In Figure~\ref{fig:supp_good_qualitative}(a), the baseline GroundNLQ fails to reason about temporal order and predicts a span before the reference span, despite feedback indicating that the target occurs after it.
Similarly, in Figure~\ref{fig:supp_good_qualitative}(d), GroundNLQ fails to incorporate the feedback and predicts the same room as the reference span, even though the feedback states that this room is incorrect.
These examples demonstrate our method's ability to effectively use user feedback.

\begin{figure*}[thbp]
    \centering
    \includegraphics[width=0.99
    \linewidth]{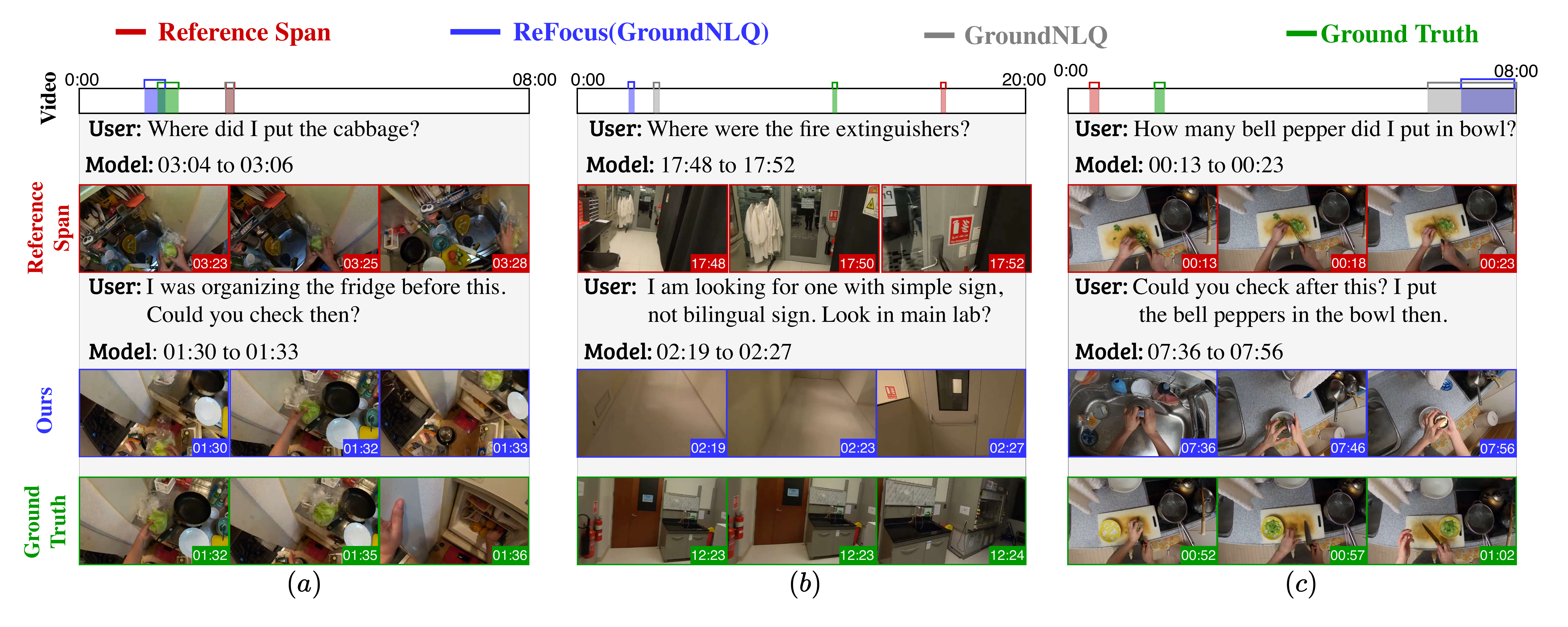}
    \caption{Additional Failure Cases. These examples show few failure cases of our method.}
    \label{fig:supp_bad_qualitative}
\end{figure*}

Figure~\ref{fig:supp_bad_qualitative} shows failure cases of our method.
In Figure~\ref{fig:supp_bad_qualitative}(a), our model uses the feedback to localize the correct moment, but the predicted span is not well aligned with the ground-truth span.
Figure~\ref{fig:supp_bad_qualitative}(b) shows a case where the model overemphasizes feedback about signs and predicts a moment containing the sign but not the fire extinguisher.
Figure~\ref{fig:supp_bad_qualitative}(c) shows a case where the model fails to identify the correct object and seems to confuse a tuna can with bell peppers.


\begin{table*}[!t]
\centering
\small
\setlength{\tabcolsep}{10pt} 
\begin{tabular}{l|cccc|cccc}
\toprule
 &
\multicolumn{4}{c|}{\textbf{\fontsize{9.5}{13}\selectfont GoalStep-QnF}} &
\multicolumn{4}{c}{\textbf{\fontsize{9.5}{13}\selectfont HD-EPIC-QnF}} \\[2pt]
 & \multicolumn{2}{c}{\textbf{IoU = 0.3}} & \multicolumn{2}{c|}{\textbf{IoU = 0.5}}
 & \multicolumn{2}{c}{\textbf{IoU = 0.3}} & \multicolumn{2}{c}{\textbf{IoU = 0.5}} \\[2pt]
\textbf{\fontsize{9.5}{13}\selectfont Method} &
\textbf{R1} & \textbf{R5} & \textbf{R1} & \textbf{R5} & \textbf{R1} & \textbf{R5} & \textbf{R1} & \textbf{R5} \\
\midrule

OSGNet & 
14.5\greentext{\textsuperscript{+0.2}} & 36.7\greentext{\textsuperscript{+0.7}} & 10.0\redtext{\textsuperscript{-0.3}} & 27.1\greentext{\textsuperscript{+0.7}} & 5.3\greentext{\textsuperscript{+0.4}} & 16.9\greentext{\textsuperscript{+0.7}} & 2.4\greentext{\textsuperscript{+0.3}} & 7.9\greentext{\textsuperscript{+0.3}} \\[2pt]

ReFocus(OSGNet) & 
\textbf{17.9\greentext{\textsuperscript{+3.6}}} & \textbf{42.0\greentext{\textsuperscript{+6.8}}} & \textbf{12.8\greentext{\textsuperscript{+2.5}}} & \textbf{32.4\greentext{\textsuperscript{+5.9}}} & \textbf{6.7\greentext{\textsuperscript{+2.2}}} & \textbf{18.6\greentext{\textsuperscript{+4.5}}} & \textbf{3.1\greentext{\textsuperscript{+0.9}}} & \textbf{8.6\greentext{\textsuperscript{+2.0}}} \\[2pt]
\midrule
GroundNLQ & 
17.7\greentext{\textsuperscript{+0.5}} & 42.2\greentext{\textsuperscript{+1.6}} & 12.5\greentext{\textsuperscript{+0.4}} & 31.4\greentext{\textsuperscript{+1.5}} & 6.6\greentext{\textsuperscript{+0.2}} & 21.3\greentext{\textsuperscript{+0.3}} & 2.9\greentext{\textsuperscript{+0.2}} & 10.7\greentext{\textsuperscript{+0.5}} \\[2pt]

ReFocus(GroundNLQ) & 
\textbf{20.7\greentext{\textsuperscript{+3.7}}} & \textbf{45.3\greentext{\textsuperscript{+5.0}}} & \textbf{14.5\greentext{\textsuperscript{+2.6}}} & \textbf{34.1\greentext{\textsuperscript{+4.3}}} & \textbf{8.2\greentext{\textsuperscript{+1.6}}} & \textbf{25.1\greentext{\textsuperscript{+4.2}}} & \textbf{3.8\greentext{\textsuperscript{+1.1}}} & \textbf{13.0\greentext{\textsuperscript{+2.6}}} \\

\bottomrule
\end{tabular}
\caption{Zero-Shot evaluation of models trained on Ego4D-QnF and tested on GoalStep-QnF and HD-EPIC-QnF. 
Deltas ($\Delta$) of model performance with feedback vs query-only are shown as superscripts.
}
\label{tab:supp_zero_shot_results}
\end{table*}

\begin{table*}[htbp]
\centering
\small
\setlength{\tabcolsep}{6.5pt} 
\begin{tabular}{l|cccc|cccc|cccc}
\toprule
 &
\multicolumn{4}{c|}{\textbf{\fontsize{9.5}{13}\selectfont Ego4D-NLQ}} &
\multicolumn{4}{c|}{\textbf{\fontsize{9.5}{13}\selectfont GoalStep-Q}} &
\multicolumn{4}{c}{\textbf{\fontsize{9.5}{13}\selectfont HD-EPIC-Q}} \\[2pt]
 & \multicolumn{2}{c}{\textbf{IoU = 0.3}} & \multicolumn{2}{c|}{\textbf{IoU = 0.5}}
 & \multicolumn{2}{c}{\textbf{IoU = 0.3}} & \multicolumn{2}{c|}{\textbf{IoU = 0.5}}
 & \multicolumn{2}{c}{\textbf{IoU = 0.3}} & \multicolumn{2}{c}{\textbf{IoU = 0.5}} \\[2pt]
\textbf{\fontsize{9.5}{13}\selectfont Method} & \textbf{R1} & \textbf{R5} & \textbf{R1} & \textbf{R5} & \textbf{R1} & \textbf{R5} & \textbf{R1} & \textbf{R5} & \textbf{R1} & \textbf{R5} & \textbf{R1} & \textbf{R5} \\
\midrule

TimeChat & 1.78 &	N/A &	0.86 &	N/A &	1.40 &	N/A &	0.40 &	N/A &	0.13 &	N/A &	0.00 &	N/A \\
UniTime & 25.04 &	N/A &	15.58 &	N/A	 &11.93 &	N/A	 &6.80 &	N/A &	5.57 &	N/A &	2.47 &	N/A  \\
\midrule
OSGNet & 29.22	& 55.89	& 20.03& 	42.57& 	29.62& 	59.24	& 24.18& 	51.84& 	14.43& 	37.77& 	9.51& 	25.14\\
ReFocus(OSGNet) & 29.26	& 54.46	& 20.49	& 42.00& 	29.95& 	58.76& 	24.70& 	51.29& 	14.51	& 37.06	& 9.59	& 24.84\\
\midrule
GroundNLQ & 29.00	& 54.92& 	21.13& 	42.35& 	23.07& 	52.87& 	17.40&	43.30 &	11.23 &	32.87 &	6.60 &	20.60 \\
ReFocus(GroundNLQ) & {29.77} &	55.03 &	{21.44} & 42.29 &	{21.90} &	{50.80} &	{16.70} &	{41.30} &	{12.10} &	{34.17} &	{7.03} &	{21.70} \\
\bottomrule
\end{tabular}
\caption{Performance comparison on NLQ-only datasets Ego4D-NLQ, GoalStep-Q, and HD-EPIC-Q.
}
\label{tab:supp_query_only}
\end{table*}

\begin{table*}[ht]
\centering
\small
\setlength{\tabcolsep}{3.2pt} 
\begin{tabular}{l|cccc|cccc|cccc}
\toprule
 &
\multicolumn{4}{c|}{\textbf{\fontsize{9.5}{13}\selectfont Ego4D-QnF}} &
\multicolumn{4}{c|}{\textbf{\fontsize{9.5}{13}\selectfont GoalStep-QnF}} &
\multicolumn{4}{c}{\textbf{\fontsize{9.5}{13}\selectfont HD-EPIC-QnF}} \\[2pt]
 & \multicolumn{2}{c}{\textbf{IoU = 0.3}} & \multicolumn{2}{c|}{\textbf{IoU = 0.5}}
 & \multicolumn{2}{c}{\textbf{IoU = 0.3}} & \multicolumn{2}{c|}{\textbf{IoU = 0.5}}
 & \multicolumn{2}{c}{\textbf{IoU = 0.3}} & \multicolumn{2}{c}{\textbf{IoU = 0.5}} \\[2pt]
\textbf{\fontsize{9.5}{13}\selectfont Method} & \textbf{R1} & \textbf{R5} & \textbf{R1} & \textbf{R5} & \textbf{R1} & \textbf{R5} & \textbf{R1} & \textbf{R5} & \textbf{R1} & \textbf{R5} & \textbf{R1} & \textbf{R5} \\
\midrule

Gemini-2.5-flash & 15.7\greentext{\textsuperscript{+1.7}} &	28.7\greentext{\textsuperscript{+0.7}} & 8.7\greentext{\textsuperscript{+2.7}} &	14.7\redtext{\textsuperscript{-1.3}} & 8.7\greentext{\textsuperscript{+2.7}} & 16.0\greentext{\textsuperscript{+1.0}} & 2.7\greentext{\textsuperscript{+0.7}} &  5.7\redtext{\textsuperscript{-1.3}} & 6.7\greentext{\textsuperscript{+4.7}} & 15.7\greentext{\textsuperscript{+6.7}} & 3.3\greentext{\textsuperscript{+3.3}} & 8.0\greentext{\textsuperscript{+4.0}} \\

\midrule
ReFocus(OSGNet)& 24.0\greentext{\textsuperscript{+3.0}} &	46.7\greentext{\textsuperscript{+2.7}} &	13.7\greentext{\textsuperscript{+2.7}} &	29.0\greentext{\textsuperscript{+3.0}} &	21.7\greentext{\textsuperscript{+2.7}} &	53.7\greentext{\textsuperscript{+3.7}} &	13.0\greentext{\textsuperscript{+3.0}} &	35.7\greentext{\textsuperscript{+2.7}} &	9.7\greentext{\textsuperscript{+1.7}} &	39.3\greentext{\textsuperscript{+2.0}} &	6.7\greentext{\textsuperscript{+1.7}} &	21.3\greentext{\textsuperscript{+1.3}}\\

{ReFocus(GroundNLQ)} & 8.7\greentext{\textsuperscript{+8.7}} & 48.0\greentext{\textsuperscript{+48.0}} & 4.7\greentext{\textsuperscript{+4.7}} & 29.3\greentext{\textsuperscript{+29.3}} & 9.7\greentext{\textsuperscript{+9.7}} & 54.7\greentext{\textsuperscript{+54.7}} &4.0\greentext{\textsuperscript{+4.0}} &38.7\greentext{\textsuperscript{+38.7}} & 4.3\greentext{\textsuperscript{+4.3}} &47.0\greentext{\textsuperscript{+47.0}} &2.0\greentext{\textsuperscript{+2.0}} &22.3\greentext{\textsuperscript{+22.3}}\\

\bottomrule
\end{tabular}
\caption{
    Performance comparison between Gemini-2.5-flash and Refocus(GroundNLQ) on a small 100 NLQ subset where ReFocus(GroundNLQ) fails with query-only but improves with feedback. 
    }
\label{tab:supp_gemini_results}
\end{table*}

\begin{table*}[ht]
\centering
\small
\setlength{\tabcolsep}{6pt} 
\begin{tabular}{l|cccc|cccc|cccc}
\toprule
 &
\multicolumn{4}{c|}{\textbf{\fontsize{9.5}{13}\selectfont Ego4D-QnF}} &
\multicolumn{4}{c|}{\textbf{\fontsize{9.5}{13}\selectfont GoalStep-QnF}} &
\multicolumn{4}{c}{\textbf{\fontsize{9.5}{13}\selectfont HD-EPIC-QnF}} \\[2pt]
 & \multicolumn{2}{c}{\textbf{IoU = 0.3}} & \multicolumn{2}{c|}{\textbf{IoU = 0.5}}
 & \multicolumn{2}{c}{\textbf{IoU = 0.3}} & \multicolumn{2}{c|}{\textbf{IoU = 0.5}}
 & \multicolumn{2}{c}{\textbf{IoU = 0.3}} & \multicolumn{2}{c}{\textbf{IoU = 0.5}} \\[2pt]
\textbf{\fontsize{9.5}{13}\selectfont Method} & $\Delta$\textbf{R1} & $\Delta$\textbf{R5} & $\Delta$\textbf{R1} & $\Delta$\textbf{R5} & $\Delta$\textbf{R1} & $\Delta$\textbf{R5} & $\Delta$\textbf{R1} & $\Delta$\textbf{R5} & $\Delta$\textbf{R1} & $\Delta$\textbf{R5} & $\Delta$\textbf{R1} & $\Delta$\textbf{R5} \\
\midrule

Gemini-2.5-flash & 8.8 & 18.6 &4.9 &	8.3 & 3.5 & 9.0 & 1.2 & 3.1 & 4.4 & 13.2 & 2.2 & 6.6 \\
{ReFocus(GroundNLQ)} & 8.3 & 47.0 & 4.9 & 25.5 & 9.8 & 51.8 & 3.9 & 34.5 & 2.9 & 45.8 & 2.2 & 22.0\\
\bottomrule
\end{tabular}
\caption{Performance comparison with feedback on subset of NLQs where Gemini-2.5-flash fails with NLQ-only.
}
\label{tab:supp_gemini_results_failed_subset}
\end{table*}

\paragraph{Zero-Shot Evaluation Results.}
In Table~\ref{tab:supp_zero_shot_results}, we report additional recall metrics not included in the main zero-shot evaluation (Table~\ref{tab:zero_shot_results}) for different methods trained only on Ego4D-QnF and evaluated on GoalStep-QnF and HD-EPIC-QnF. 
We observe that \modelname~with GroundNLQ outperforms all other methods across all metrics and datasets in the zero-shot setting. 
As in the main results, \modelname~enables more effective use of user feedback, leading to larger gains across all recall metrics than the baseline without \modelname, even in the zero-shot setting.

\paragraph{Query Only Performance.}
In Table~\ref{tab:supp_query_only}, we show the query-only performance of our approach compared to other baselines. 
For both GroundNLQ and OSGNet, we see similar performance with and without our approach \modelname.
In other words, our module (FALM) does not significantly change the performance of the base model when feedback is not available, which is desirable as we want to maintain the original performance of the base model on query-only cases.
For both EM-NLQ models, our approach lead to significant improvements when feedback is available, as shown in Table~\ref{tab:combined_feedback_results}.

\subsection{Commercial LVLM Evaluation}
\label{ssec:supp_commercial_lvlm}
We evaluate Gemini-2.5-flash to assess how effectively a state-of-the-art commercial LVLM can incorporate user feedback.
In Table~\ref{tab:gemini_results}, we evaluate Gemini on a subset of queries for which GroundNLQ fails completely and report results on two of the three QnF datasets.
Here, we provide the full table for the same evaluation in Table~\ref{tab:supp_gemini_results}.

We also present an additional evaluation in Table~\ref{tab:supp_gemini_results_failed_subset} on queries for which Gemini itself fails.
Specifically, we first evaluate Gemini on a set of EM-NLQs and then select the examples for which it fails to localize the correct response span within its top-5 predictions using only the query.
We then evaluate Gemini on these examples using user feedback from our datasets.
The resulting subset contains 204, 255, and 273 feedback instances for Ego4D-QnF, GoalStep-QnF, and HD-EPIC-QnF, respectively, out of 300 total feedback instances per dataset.

Table~\ref{tab:supp_gemini_results_failed_subset} shows the performance gains from adding user feedback over the query-only setting for Gemini-2.5-flash and our approach on this subset.
While Gemini-2.5-flash shows gains similar to our model in Recall@1 at IoU$=0.3$ on Ego4D-QnF and HD-EPIC-QnF, our approach performs substantially better at Recall@5 and consistently better across all metrics on GoalStep-QnF.

Comparing the gains in Table~\ref{tab:gemini_results} with those in this new evaluation, we find that Gemini-2.5-flash improves less in Table~\ref{tab:gemini_results}.
This suggests that Gemini-2.5-flash can be confused by user feedback in cases where it can already localize the correct response span from the NLQ alone.

We also provide the prompts used with Gemini-2.5-flash for both the NLQ-only and NLQ-with-feedback settings in Figure~\ref{fig:supp_gemini_prompts}.

\subsection{Limitations and Future Work} 
\label{ssec:supp_limitations_future_work}

Collecting user feedback for the EM-NLQ task is costly. 
In this work, we propose an effective method for synthesizing feedback and training models for interactive EM-NLQ, yielding improved performance.
However, the approach relies on LLMs to generate feedback, and some generated examples contain hallucinations that introduce noise into the training data.
Reducing these hallucinations through stronger LLMs, better system design, or improved prompts could produce cleaner training data and further improve performance.

Additionally, we generate feedback based on language descriptions of the reference and target spans.
Incorporating visual context from the corresponding video spans could enable more precise and visually grounded feedback, potentially improving feedback quality.
However, this would substantially increase computational cost because of the larger context required to process video input.

Furthermore, although we show that our method can handle multiple rounds of feedback by modeling them independently and combining them to form a strong baseline, we do not extensively study multi-turn user feedback in which later feedback may refine, correct, or depend on earlier feedback.
Training a specialized method with order-aware state and belief tracking for multiple rounds of feedback could further improve performance.
Future work could explore such approaches to build a more robust conversational system.

\subsection{Human User Feedback Collection}
\label{ssec:supp_human_user_feedback_collection}
\begin{figure*}
    \centering
    \includegraphics[width=0.97\linewidth]{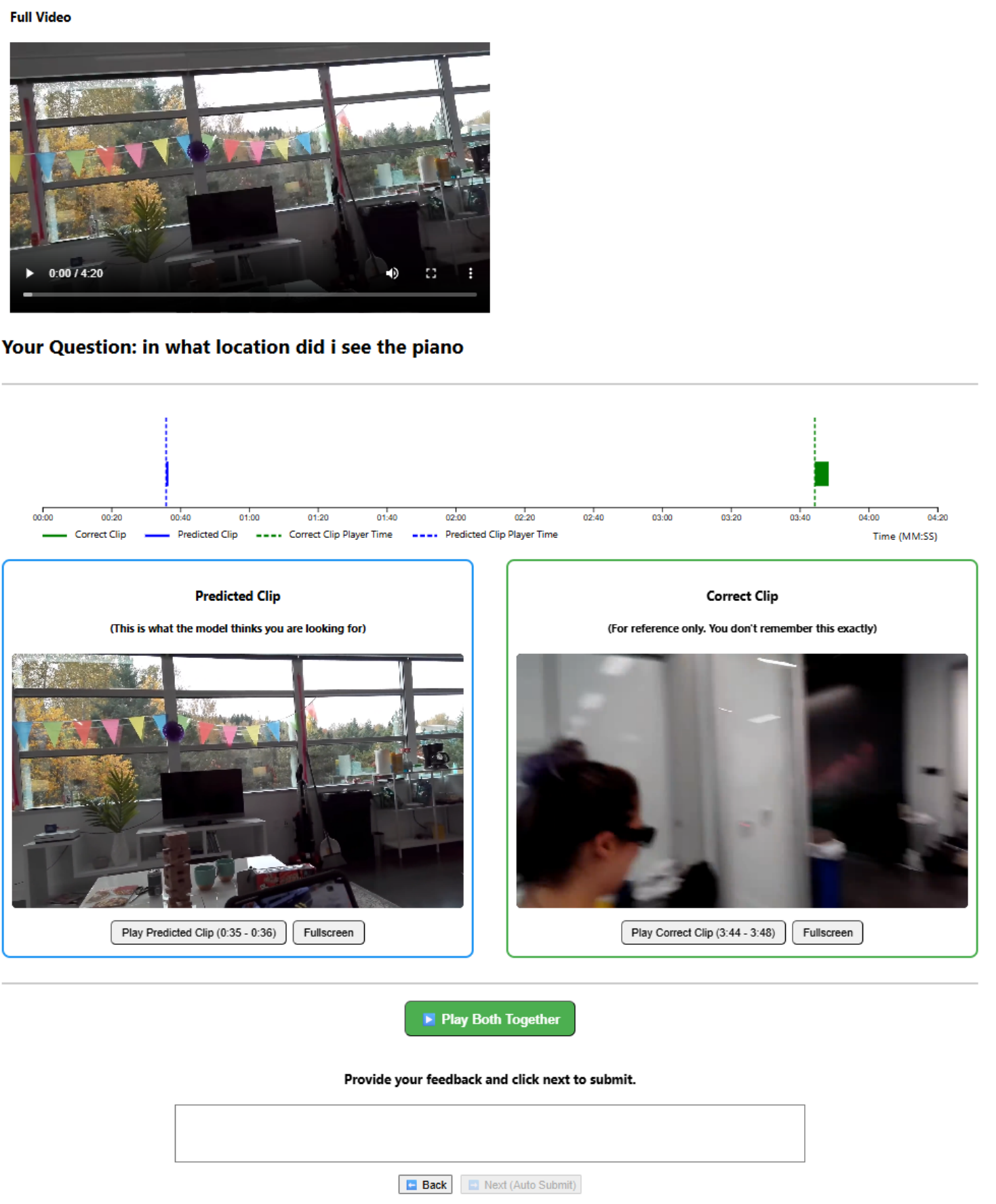}
    \caption{Human User Feedback Collection UI. Users were given a short tutorial and advised not to directly answer the question.
    }
    \label{fig:feedback_collection}
\end{figure*}

We collect human feedback from 11 users, who are asked to assume the role of a user issuing a query in the EM-NLQ task to recall a past activity.
Given the query, the full video, and an incorrect reference span, the annotators are instructed to provide feedback that helps identify the correct response span.
We instruct annotators not to answer the query directly by referring to the correct response span and to avoid relying only on simple temporal feedback, such as searching before or after the current reference span.
Figure~\ref{fig:feedback_collection} shows the interface used to collect human user feedback.

\subsection{Query Generation for GoalStep and HD-EPIC}
\label{ssec:supp_nlq_generation}
To evaluate our approach across diverse scenarios and activity types in egocentric video, we extend two popular datasets, Ego4D-GoalStep~\cite{goalstep} and HD-EPIC~\cite{hdepic}, to be compatible with the EM-NLQ task and with our feedback-based setup.
While Ego4D-NLQ~\cite{ego4d} covers a wide range of daily activities, Ego4D-GoalStep contains structured, multi-step tasks driven by explicit goals, and HD-EPIC focuses on kitchen-centered activities. 
Ego4D-GoalStep and HD-EPIC provide rich annotations, but they do not include EM-NLQ style natural language queries or feedback. 
Therefore, first we augment these datasets with queries, then along with Ego4D, with feedback annotations such that we enable controlled evaluation on theses diverse cases for our model and the baselines. 
Next, we describe the query generation process for Ego4DGoalStep and HD-EPIC.

\begin{figure*}[thbp]
\centering
    \includegraphics[width=0.99\linewidth]{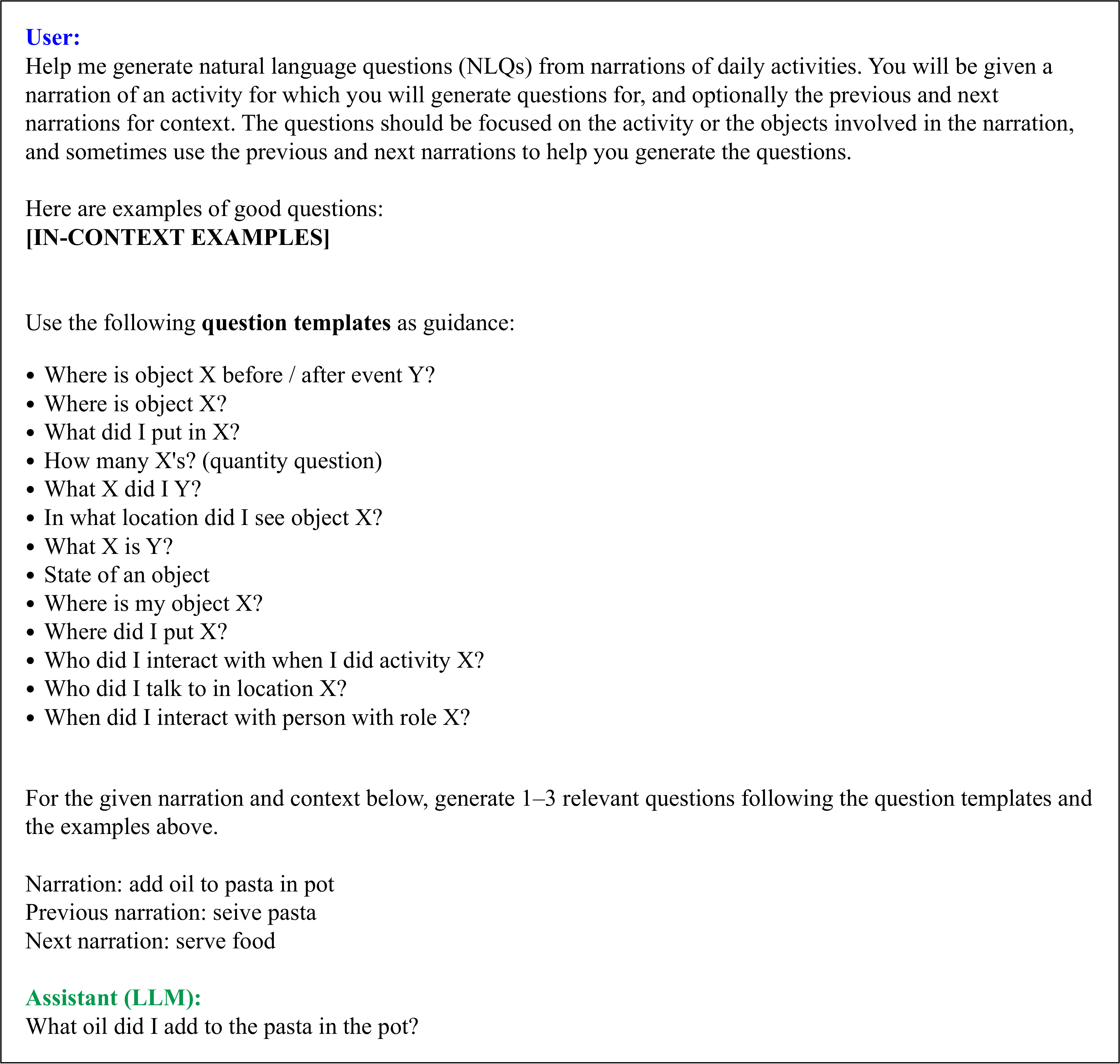}
\caption{Example prompt used to generate GoalStep-Q and HD-EPIC-Q datasets and output generated by Qwen-3-8B. Text after Assistant is generated output. We provide 8 in-context examples.
}
\label{fig:supp_nlq_gen}
\end{figure*}

For these datasets we leverage the provided text-based steps or narrations annotations. 
That is, we have a given set of text annotations $\{n_1, n_2,...,n_m\}$ and their corresponding video spans $\{s_1,s_2,...,s_m\}$ sorted by their timestamps for a egocentric video $\mathcal{V}$. 
Then, for each narration $n_x$, we collect $q$ textual NLQs $\mathcal{Q}_{x} = \{Q_{x,1}, Q_{x,2}, ..., Q_{x,q}\}$ with response span $s_x$ by prompting a Large Language Model (LLM), Qwen-3-8B~\cite{qwen3}, with previous, current, and next narration i.e. $\mathcal{Q}_{n_x}=\text{LLM}(n_{x-1},n_x,n_{x+1})$. 
Additionally, we provide the LLM with the NLQ templates used in Ego4D-NLQ to collect annotations along with 8 in-context examples.
Figure \ref{fig:supp_nlq_gen} shows an example showcasing the prompt used and the generated output.

We refer to the query-extended version of these datasets as GoalStep-Q and HD-EPIC-Q, and Table \ref{tab:supp_nlq_dataset_statistics} shows the statistics of these datasets along with Ego4D-NLQ across multiple splits. 
On average, we notice that these datasets have longer video duration compared to Ego4D-NLQ, and also have very different response span durations, allowing us to test the models on diverse scenarios.

\begin{table}[t]
\centering
\small
\setlength{\tabcolsep}{3pt} 
\begin{tabular}{lccccc}
\hline
\textbf{Dataset} & \textbf{Split} & \textbf{\#}$\mathcal{Q}$ & \textbf{\#}$\mathcal{V}$ & $\mathcal{V}$ Dur.(s) & $\mathcal{R}^q$ Dur.(s) \\
\hline

\multirow{2}{*}{Ego4D-NLQ} & Train & 13847 & 1271 & 529 & 11.3 \\
& Val & 4552 & 415 & 548 & 10.75 \\
\midrule

\multirow{3}{*}{GoalStep-Q} & Train & 13849 & 606 & 1014 & 31.6 \\
& Val & 1554 & 35 & 960 & 30.8 \\
& Test & 3000 & 35 & 949 & 29.9 \\
\midrule

\multirow{3}{*}{HD-EPIC-Q}  & Train & 13849 & 17 & 938 & 2.0 \\
& Val & 1554 & 15 & 1053 & 2.2 \\
& Test & 3000 & 15 & 1061 & 2.2 \\

\bottomrule
\end{tabular}
\caption{Summary of EM-NLQ datasets characteristics for all splits, reporting the total number of queries $\mathcal{Q}$, number of videos $\mathcal{V}$, average video duration, and average response span $\mathcal{R}$ duration in seconds.}
\label{tab:supp_nlq_dataset_statistics}
\end{table}

\subsection{Feedback Generation for All Datasets} %
\label{ssec:supp_qnf_datasets}
In this section, we provide more details on the feedback generation process based on the recipe from Sec.~\ref{ssec:recipe-data}. %
For an egocentric video $\mathcal{V}$, a query $\mathcal{Q}$, and a ground-truth response span $\mathcal{R}^q$, we sample a reference span $\mathcal{R}^f$ and caption both $\mathcal{R}^q$ and $\mathcal{R}^f$ to get textual descriptions $\mathcal{D}^q$ and $\mathcal{D}^f$, respectively. 
We then generate an explanation $E^q$ from $\mathcal{Q}$ and $\mathcal{R}^q$. 
Finally, we generate user feedback $\mathcal{F}$ using $\mathcal{D}^q$, $\mathcal{D}^f$, $E^q$, and the relative temporal ordering between $\mathcal{R}^q$ and $\mathcal{R}^f$. 
Next, we describe each step in more detail.

\paragraph{Reference Span Sampling.} 
We sample two different types of reference spans: query-relevant and query-irrelevant spans. 
We describe how we sample each of these type of reference spans below:

\textit{\textbf{1)} Query-Irrelevant Reference Spans}: For a query $\mathcal{Q}$ with ground-truth response span $\mathcal{R}^q$, we sample a random span $\mathcal{R}^f$ such that the temporal intersection-over-union satisfies $\textbf{\textit{tIoU}}(\mathcal{R}^q,\mathcal{R}^f)=0$. 
Random spans are generated by sampling a center timestamp and a span duration. 
The center timestamp is sampled uniformly from the egocentric video $\mathcal{V}$, while the span duration is sampled from a beta distribution fitted to the min-max normalized response span durations in the Ego4D-NLQ training set.

Additionally, for other queries $\mathcal{Q}' \neq \mathcal{Q}$ in the same video $\mathcal{V}$, we also consider their response spans $\mathcal{R'}^f$ as query-irrelevant reference spans for $\mathcal{Q}$ when $\textbf{\textit{tIoU}}(\mathcal{R}^q,\mathcal{R'}^f)=0$.

\textit{\textbf{2)} Query-Relevant Reference Spans}: We sample a reference span $\hat{\mathcal{R}}^f$ from the outputs of a pretrained EM-NLQ model. 
Specifically, we select the top-1 prediction $\hat{\mathcal{R}}^q_1$ from the GroundNLQ~\cite{groundnlq} model pretrained on NaQ+NLQ~\cite{naq} for queries where Recall@5, IoU=0.3 is 0.

Furthermore, we also sample reference spans that are visually similar to the ground-truth span of $\mathcal{Q}$ but do not intersect with it. 
To do this, we create a set of candidate reference spans using a sliding-window procedure, where each span $\mathcal{R'}{i}$ contains the same number of $d$ frames as $\mathcal{R}^q$ and uses a stride of $\lfloor d/4 \rfloor$, resulting in $R={\mathcal{R'}1,\mathcal{R'}2,\ldots,\mathcal{R'}m}$, where $m$ is the total number of spans. 
We embed each candidate span using EgoVideo's ViT-1B~\cite{egovideo}, averaging clip embeddings when $d$ exceeds the video encoder's frame size, to obtain $E(R)={f_1,f_2,\ldots,f_m}$, where $f_i$ denotes the embedding of $\mathcal{R'}{i}$. 
Finally, we select $\mathcal{R'}{i^*}$ such that $\textbf{\textit{tIoU}}(\mathcal{R}^q,\mathcal{R'}{i^*})=0$ and $\cos(f_r,f{i^*})$ is maximized, where $f_r$ is the embedding of the ground-truth response span $\mathcal{R}^q$ and $\cos(\cdot,\cdot)$ denotes cosine similarity.

\begin{figure*}
    \centering
    \includegraphics[height=0.96\textheight]{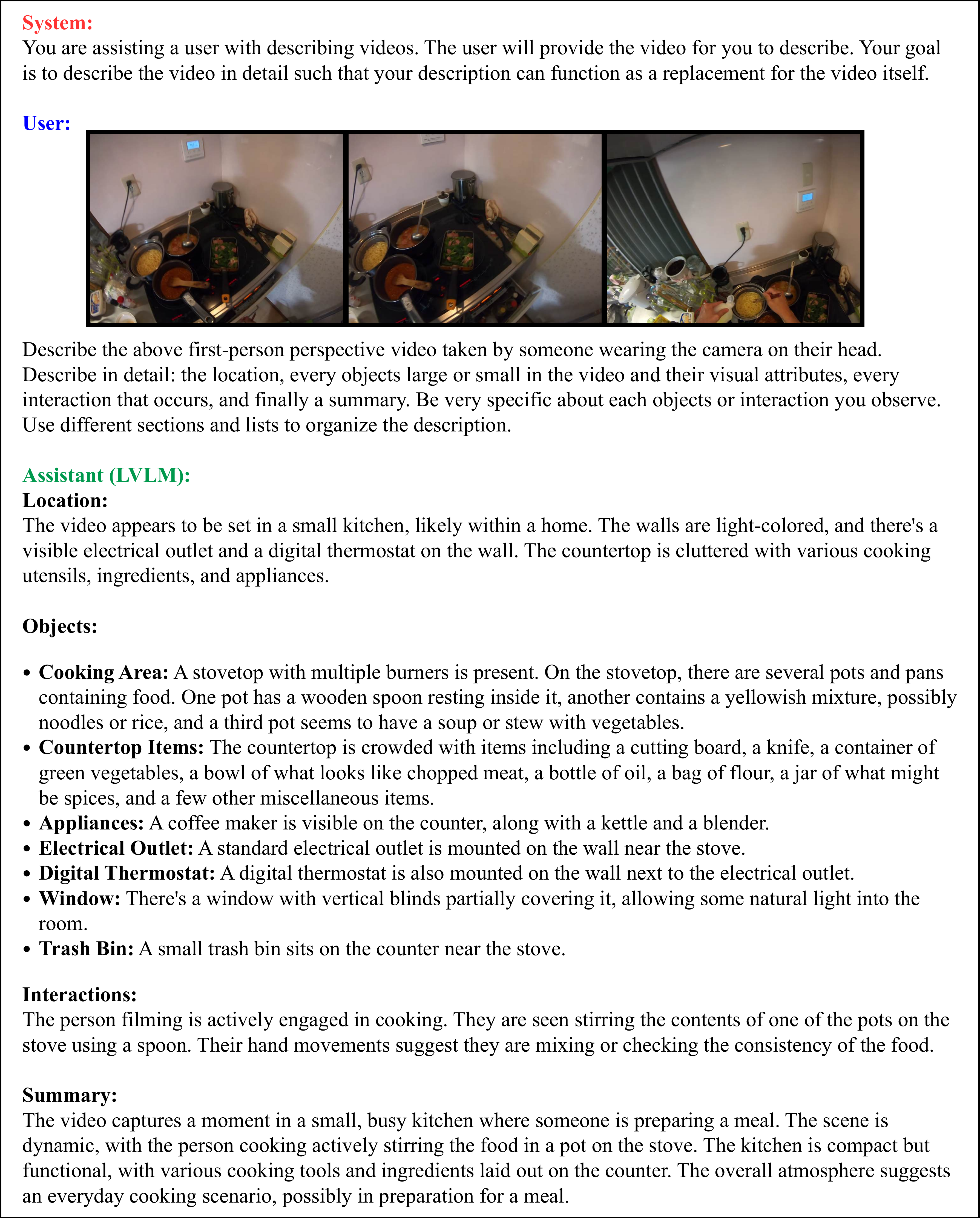}
    \caption{Example Span Captioning prompt and output generated by Qwen-2.5-VL-Instruct during user feedback generation.
    }
    \label{fig:supp_span_caption}
\end{figure*}

\paragraph{Span Captioning.} For all response spans $\mathcal{R}^q_i$ and $\mathcal{R}^f_j$, we use Qwen-2.5-VL-7B-Instruct~\cite{qwen25vl} to generate captions $\mathcal{D}^q_i$ and $\mathcal{D}^f_j$ that describe the spans. 
We resize the video so that its shorter side is 480 px while preserving the aspect ratio. 
We then sample frames at 3 frames per second, up to a maximum of 96 frames. 
For longer spans, we uniformly sample 96 frames instead. 
Figure~\ref{fig:supp_span_caption} shows the prompt and an example output for response span captioning. 
The generated descriptions include detailed information about the video content, including the scene type, relevant objects and their attributes, and the interactions.

\begin{figure*}
    \centering
    \includegraphics[width=0.99\linewidth]{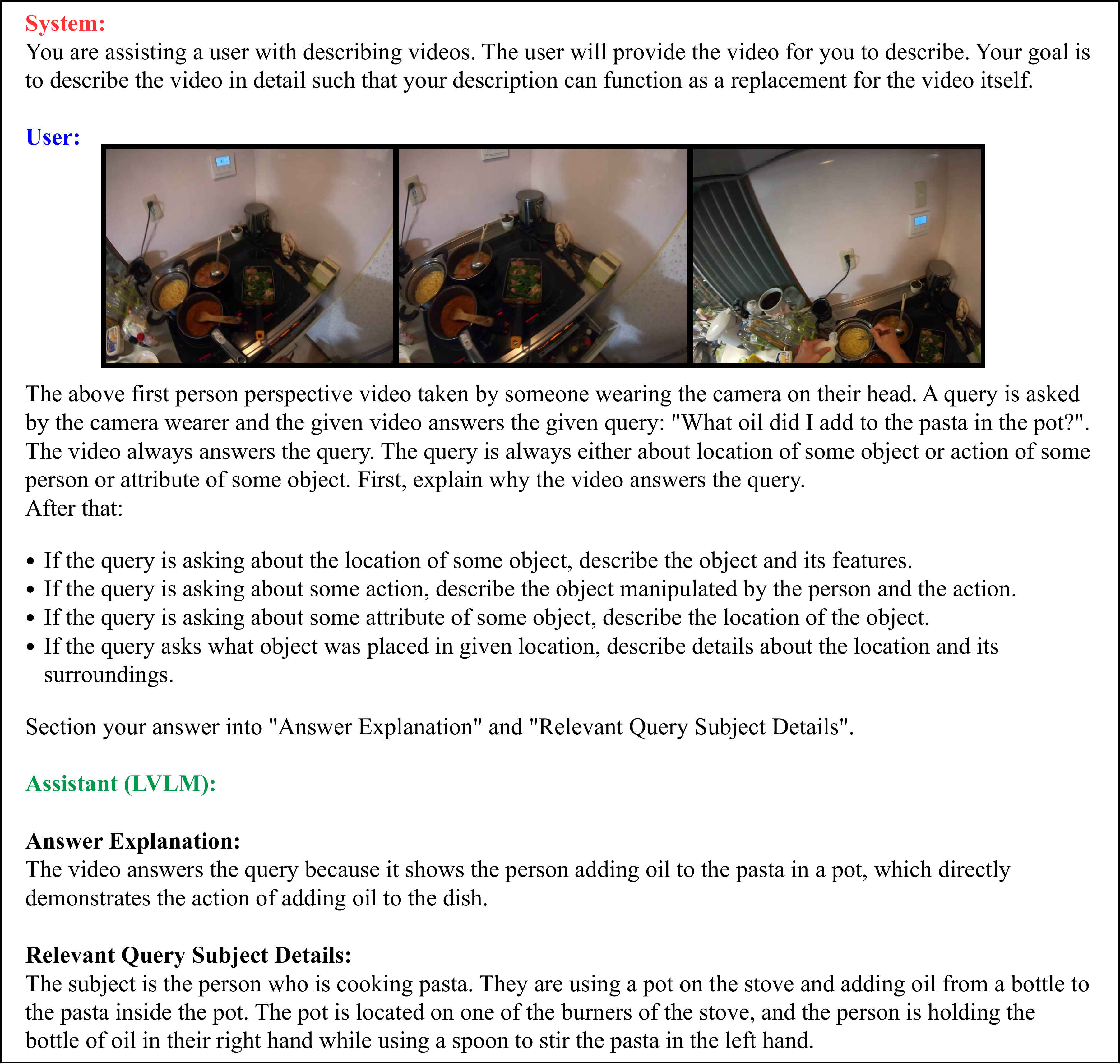}
    \caption{Example Response Span Explanation prompt and output generated by Qwen-2.5-VL-Instruct during user feedback generation.
    }
    \label{fig:supp_span_expl}
\end{figure*}

\paragraph{Response Span Explanation.} Similar to span captioning, we use Qwen-2.5-VL-7B-Instruct~\cite{qwen25vl} to generate an explanation $E^q_i$ from $\mathcal{R}^q_i$ and $\mathcal{Q}_i$. 
These explanations add specific details about why the response span answers the given query and are ultimately used during feedback generation to avoid answering the query directly. 
We resize the video and sample frames in the same way as for span captioning. 
Figure~\ref{fig:supp_span_expl} shows the prompt and an example explanation for a response span.

\begin{table}[tb]
\centering
\small
\setlength{\tabcolsep}{3.5pt} 
\begin{tabular}{lcccccc}
\hline
\textbf{Dataset} & \textbf{Split} & \textbf{\#}$\mathcal{F}$ & \textbf{\#}$(\mathcal{R}^q,\mathcal{R}^f)$ & \textbf{\#}$\mathcal{F}/Q$  & \textbf{\#}$\mathcal{F}_{\text{rel}}$ \\
\midrule
\multirow{2}{*}{Ego4D-QnF} 
 & Train & 183490  & 46433 & 13.46 & 17739 \\
 & Val   & 22071 & 14122 & 5.0 &  1913 \\
\midrule
\multirow{3}{*}{GoalStep-QnF} 
 & Train & 131405 & 32829 & 9.51 & 17988\\
 & Val   & 7335 & 4691 & 5.0 & 594 \\
 & Test   & 14083 & 9038 & 5.0 & 1178 \\
\midrule
\multirow{3}{*}{HD-EPIC-QnF} 
 & Train & 117016 & 29801 &  9.14 & 16508 \\
 & Val   & 7770 & 4973 &   5.0 & 630\\
 & Test   & 21418 & 9623 &   5.0 & 1258 \\
\hline
\end{tabular}
\caption{Statistics of the feedback datasets. For each split, we report the number of feedback (\#\(\mathcal{F}\)), the number of response span and reference span pairs from which feedbacks were generated as \#(\(\mathcal{R}^q,\mathcal{R}^f\)), the average number of feedback per query (\#\(\mathcal{F}/Q\)), and the number of feedback from query-relevant reference spans (\#\(\mathcal{F}_{\text{rel}}\)).}
\label{tab:supp_feedback_datasets}
\end{table}

\paragraph{Feedback Generation.}
Finally, we use the generated $\mathcal{D}^q_i$, $\mathcal{D}^f_j$, and $E^q_i$ to generate user feedback. 
In addition to these captions, we provide the relative temporal ordering between $\mathcal{R}^q_i$ and $\mathcal{R}^f_j$ to a reasoning LLM, Qwen-QwQ-32B-AWQ~\cite{qwen-qwq}, which produces feedback $\mathcal{F}_{i,j}$. 
We use a reasoning LLM because generating effective user feedback requires non-trivial reasoning.

\begin{figure*}
    \centering
    \includegraphics[width=0.97\linewidth]{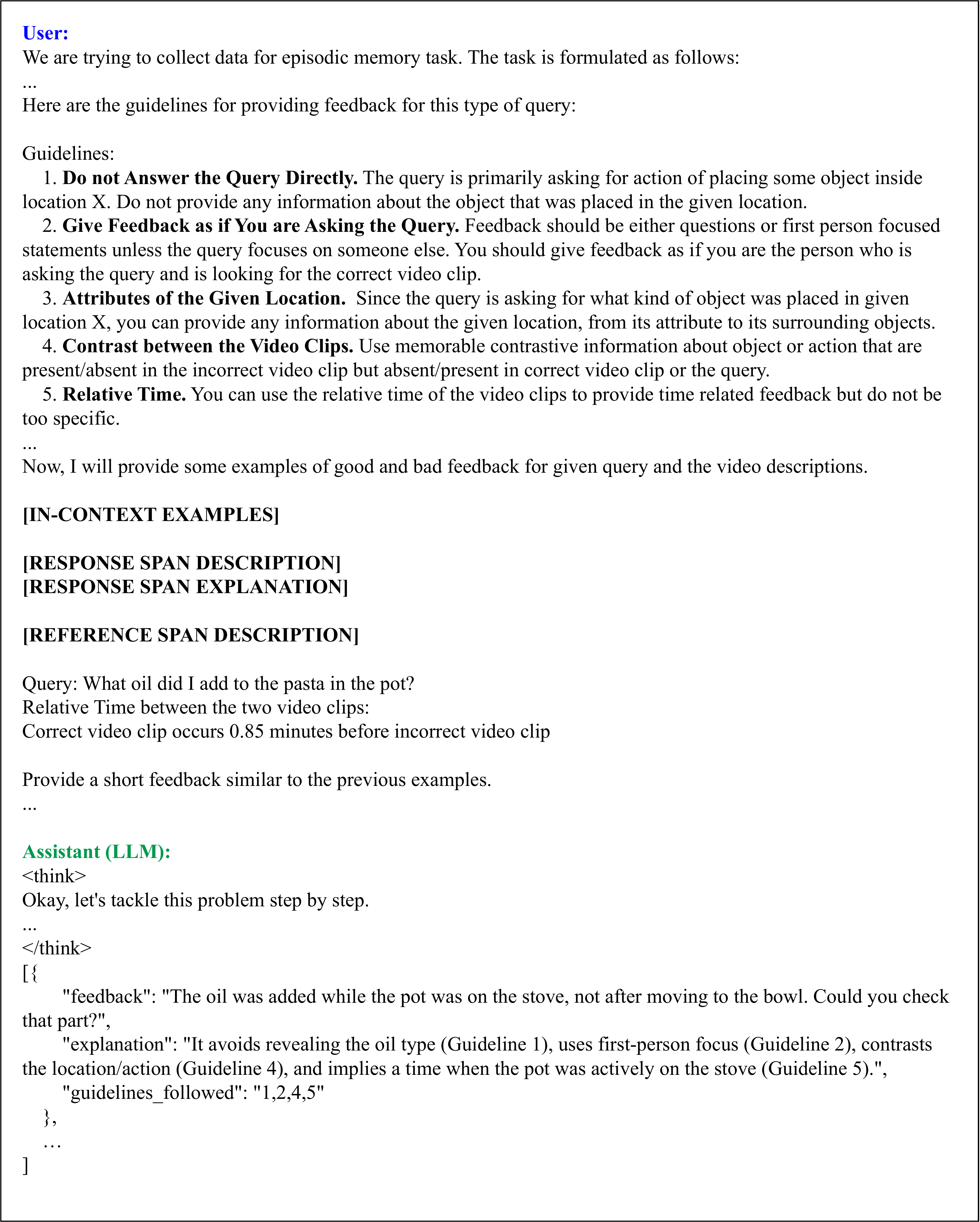}
    \caption{Example Feedback Generation prompt and output generated by Qwen-QwQ-32B-AWQ during user feedback generation. Prompt and output are truncated for brevity.
    }
    \label{fig:supp_feedback_gen}
\end{figure*}

For feedback generation, we sample five in-context examples, prioritizing examples that share the same query template as the current query. 
We prompt the model to add more detail about the query subject, contrast the reference span with the target span, use the relative temporal order, or combine these cues to produce feedback. 
Figure~\ref{fig:supp_feedback_gen} shows a truncated example prompt for feedback generation and its output.

\paragraph{Simple Temporal Feedback.}
In addition to the previous feedback, we also generate simple temporal feedback, i.e., feedback that only indicates whether the model should search \emph{before} or \emph{after} the current reference span. 
This type of feedback doesn't require complex reasoning and can be generated with simple templates based on the relative temporal order between the response span and the reference span.
We sample this temporal user feedback from a predefined set of feedback statements such as \emph{`I think it was before this'}, \emph{`After this moment'}, or \emph{`look before this'}. 
We also use this type of feedback during training as a form of data augmentation.

\paragraph{Dataset Generation and Statistics.} 

We sample 6 random reference spans per video $\mathcal{V}_i$ in the datasets. 
We additionally sample 5000 and 1000 query-relevant spans for the Ego4D-NLQ training and validation splits\footnote{Ego4D-NLQ test set is private.}, respectively. 
For GoalStep-Q and HD-EPIC-Q, we sample 5000, 333, and 666 query-relevant spans for the training, validation, and test splits, respectively. 
For each query in Ego4D-NLQ, we sample 3 reference spans to generate feedback, while we sample 2 for GoalStep-Q and HD-EPIC-Q.

In all datasets and for each query $\mathcal{Q}_i$ in the validation and test splits, we sample five feedback instances: two from query-relevant spans when available, two from query-irrelevant spans, and one simple temporal feedback instance. 
Table~\ref{tab:supp_feedback_datasets} reports the statistics of the generated Ego4D-QnF, GoalStep-QnF, and HD-EPIC-QnF datasets.

\begin{figure*}
    \centering
    \includegraphics[width=0.97\linewidth]{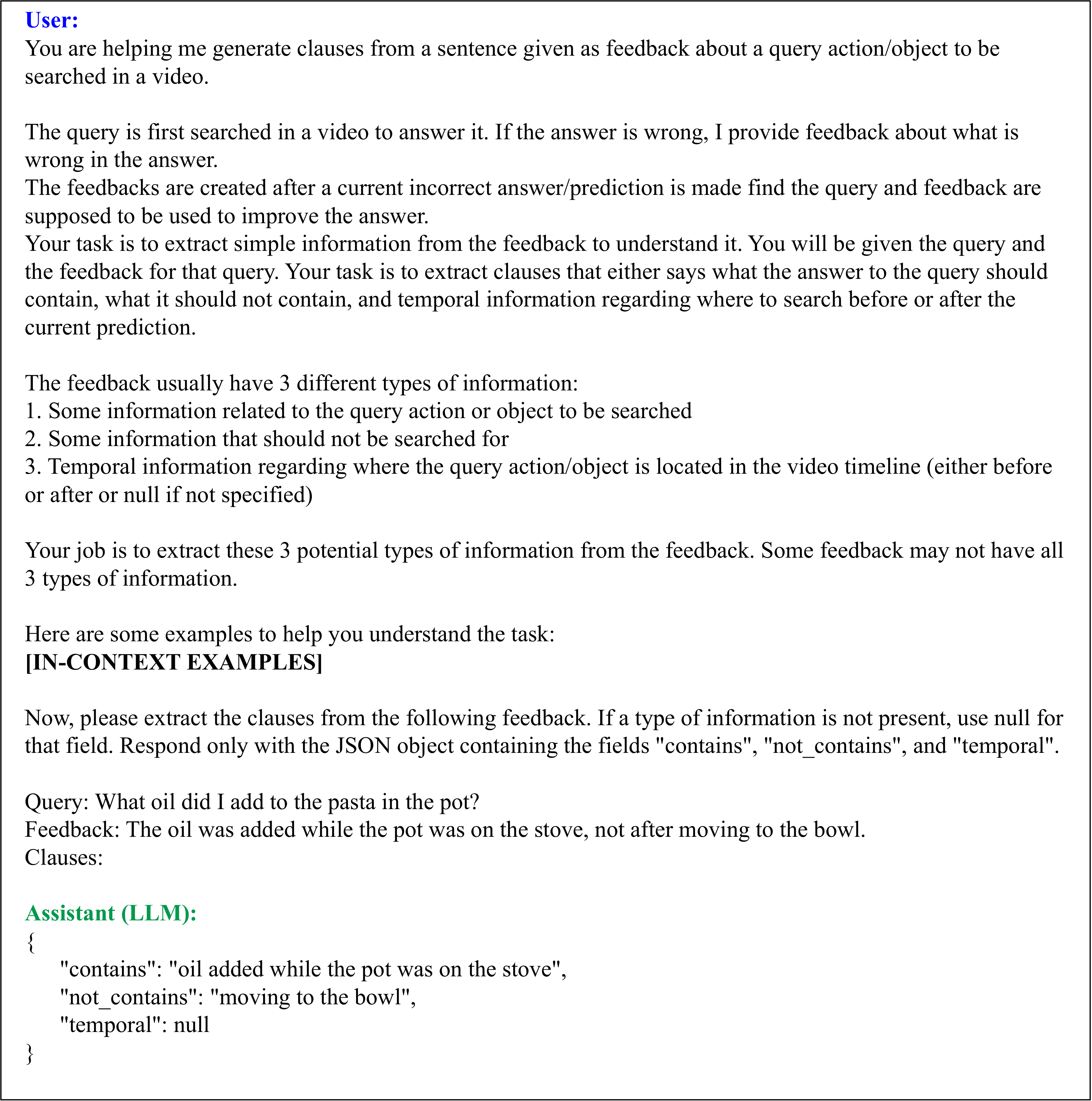}
    \caption{Example Clauses Extraction prompt and output generated by Qwen3-8B.
    }
    \label{fig:supp_clauses_extraction}
\end{figure*}

\paragraph{Clause Extraction from Feedback.}
As described in Sec.~\ref{ssec:pnp-module}, we use an LLM to extract 3 types of information: what the response span should contain, what it should not contain, and whether to search content before or after the current reference span.
We use Qwen3-8B with in-context examples to extract the \textit{contains}, \textit{not-contains}, and \textit{temporal} clauses.
Given the query $\mathcal{Q}_i$ and its corresponding feedback $\mathcal{F}_{i,x}$, we prompt the LLM to decompose the feedback into these 3 clause types when they are present.
Figure~\ref{fig:supp_clauses_extraction} shows the prompt and an example output.

\subsection{Evaluation Setup}
\label{ssec:supp_eval_setup}
In this section, we provide additional details about the methods used in our experiments.

\begin{figure*}[t]
    \centering

    \begin{subfigure}[t]{0.97\linewidth}
        \centering
        \includegraphics[width=\linewidth]{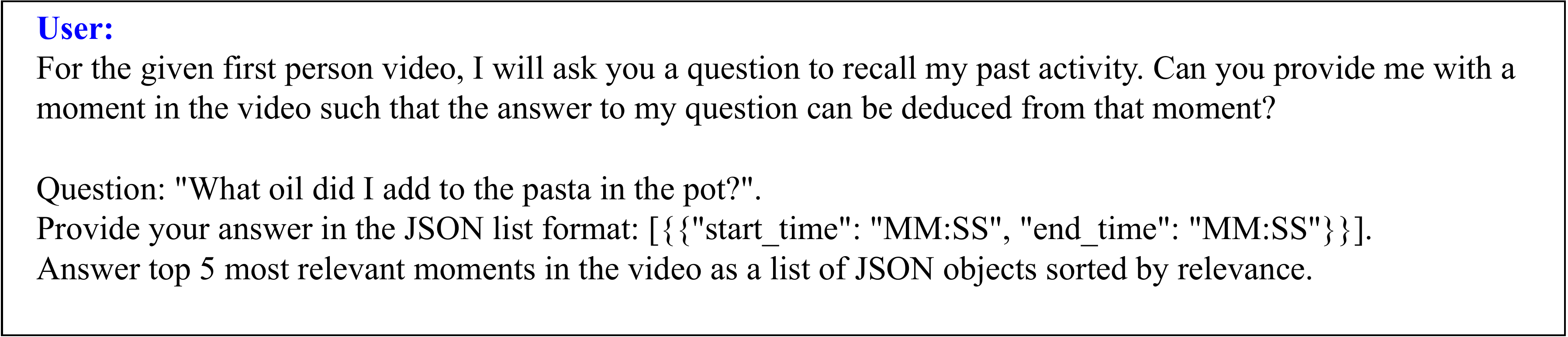} 
        \label{fig:supp_gemini_prompt1}
         \vspace{-0.2cm}
        \caption{}
    \end{subfigure}%
    \hfill
    \begin{subfigure}[t]{0.97\linewidth}
        \centering
        \includegraphics[width=\linewidth]{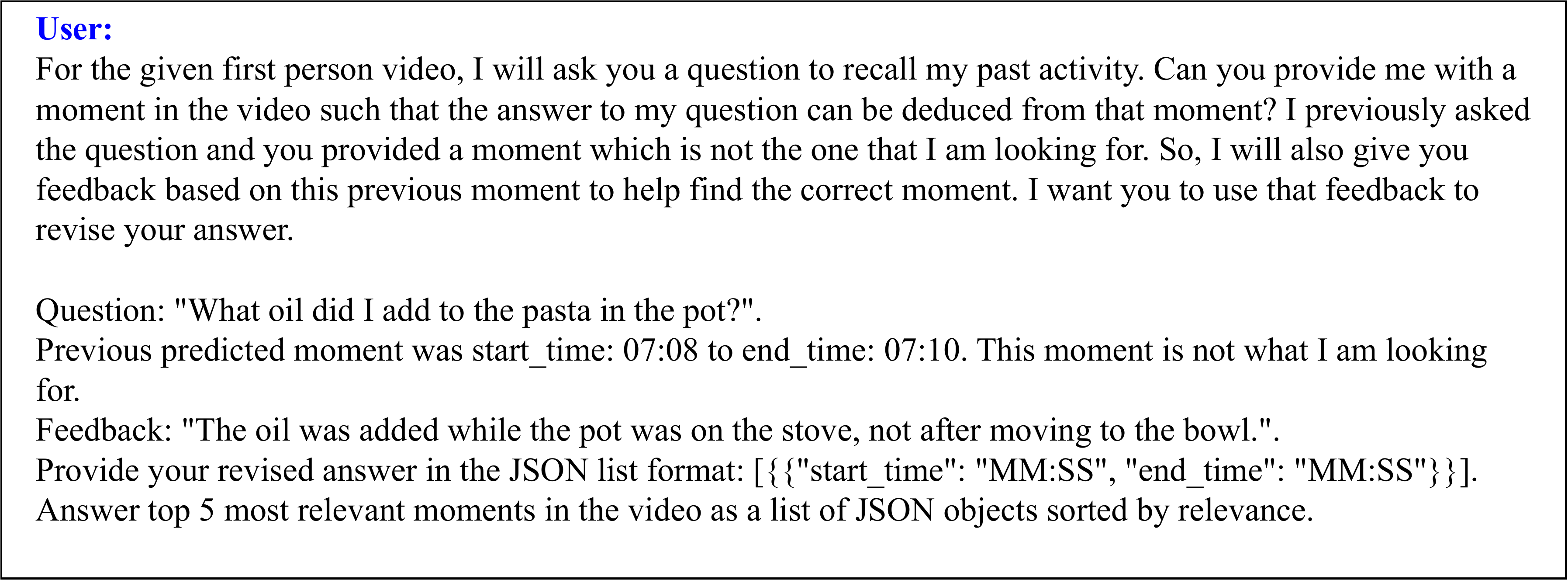} 
        \label{fig:supp_gemini_prompt2}
        
     \vspace{-0.4cm}
        \caption{}
    \end{subfigure}

    \caption{
       Example prompts used with Gemini-2.5-flash. (a) shows inference with NLQ only. (b) shows inference with NLQ and user feedback.
    }
    \label{fig:supp_gemini_prompts}
    \vspace{-0.5cm}
\end{figure*}

\textbf{TimeChat} \cite{timechat}: TimeChat uses video Q-Former along with timestamp aware frame encoders to align visual context with temporal information. 
TimeChat finetunes LLaMA-2~\cite{llama2} using LoRA~\cite{lora} on the TimeIT dataset on multiple tasks that require fine-grained temporal understanding. 
We modify the existing temporal video grounding prompt slightly to add the reference span and user feedback information similar to the Gemini-2.5-flash prompt in Figure \ref{fig:supp_gemini_prompts}.
We evaluate TimeChat without any training on user feedback (zero-shot evaluation). 

\textbf{UniTime} \cite{unitime}: UniTime leverages dynamic scaling o spatial features, interleaves timestamp information with frames, and hierarchical inference strategy to handle long videos. 
UniTime finetunes Qwen2-VL~\cite{qwen2-vl} using LoRA on several temporal localization datasets notably NaQ \cite{naq} and Ego4D-NLQ \cite{ego4d} resulting in the best performance on Ego4D-NLQ among LVLM methods. 
We evaluate UniTime in both zero-shot and fine-tuned settings. %
To adapt UniTime to user feedback, we use a prompt that includes the reference span information along with the user feedback similar to the Gemini-2.5-flash prompt in Figure \ref{fig:supp_gemini_prompts}. 
For the finetuned setting, we finetune on all 3 EM-QnF datasets simultaneously. 
We do so by sampling equal number of query and query with feedback samples for 1 epoch.

\textbf{GroundNLQ} \cite{groundnlq} GroundNLQ uses multi-scale modules to constructs the text-aware video feature pyramid and lightweight decoders to find and estimate moment boundaries. 
In our experiments, we use global attention instead of local attention. 
To adapt the base model to feedback input, we simply concatenate the text features with prediction span and feedback features to form a new query representation. 
We use a GroundNLQ model pretrained on NaQ+NLQ~\cite{naq} dataset and finetune with a learning rate of $10^{-6}$ and a batch size of 16.

\textbf{OSGNet} \cite{osgnet} OSGNet is a state-of-the-art approach that integrates object features from CoDETR~\cite{codetr} to capture object information not represented with video features. 
It also uses shot level contrastive learning to understand the camera wearer's attention from frequent movements inherent to egocentric videos. 
For our experiments, we use a OSGNet model pretrained on NaQ+NLQ and finetune on EM-QnF datasets. 
We follow the same concatenation with prediction span and feedback features to form a new query to adapt OSGNet to feedback data. 
We disable shot-level contrastive loss and finetune with a learning rate of $10^{-4}$ and a batch size of 4.

\paragraph{Finetuning with User Feedback.} We finetune all EM-NLQ models on both NLQ-only and user feedback data simultaneously. 
For all experiments, we randomly sample equal number of NLQ-only samples and NLQ with user feedback samples per epoch. 
So across many optimization steps, each method is trained on equal share of NLQ-only and user feedback data from each of the datasets.

We also generate simple temporal user feedback described in Sec \ref{ssec:supp_qnf_datasets} during finetuning. 
Each epoch, we generate $20\%$ of existing feedback data as these simple temporal feedback.

\end{document}